\documentclass{article}

\usepackage{arxiv}

\usepackage[utf8]{inputenc} 
\usepackage[T1]{fontenc}    
\usepackage{hyperref}       
\usepackage{url}            
\usepackage{booktabs}       
\usepackage{amsfonts}       
\usepackage{nicefrac}       
\usepackage{microtype}      
\usepackage{lipsum}
\usepackage{graphicx}
\usepackage{graphicx}
\usepackage{color}
\usepackage{amsmath}
\usepackage{amsfonts}
\usepackage{adjustbox}
\usepackage{multirow}
\graphicspath{ {./images/} }

\newcommand{\Ep}[2]{\mathbb{E}_{#1}\left[{#2}\right]}

\newcommand{\ignore}[1]{}

\newcommand{\KL}[2]{D_{\text{KL}} \left( #1 \middle\| #2 \right)}

\title{Bayesian Neural Networks at Scale: A Performance Analysis and Pruning Study}

\author{
 Himanshu Sharma \thanks{corresponding author} \\
  Argonne Leadership Computing Facility \\
  Argonne National Laboratories\\
  Lemont, IL, 60439  \\
  \texttt{himanshu90sharma@gmail.com} \\
  \And
  Elise Jennings \\
  Argonne Leadership Computing Facility \\
  Argonne National Laboratories\\
  Lemont, IL, 60439  \\
}

\begin{document}
\maketitle
\begin{abstract}
Bayesian neural Networks (BNNs) are a promising method of obtaining statistical uncertainties for neural network predictions but with a higher computational overhead which can limit their practical usage. This work explores the use of high performance computing with distributed training to address the challenges of training BNNs at scale. We present a performance and scalability comparison of training the VGG-16 and  Resnet-18 models on a Cray-XC40 cluster. We demonstrate that network pruning can speed up inference without accuracy loss and provide an open source software package, {\it{BPrune}} to automate this pruning. For certain models we find that pruning up to 80\% of the network results in only a 7.0\% loss in accuracy.  With the development of new hardware accelerators for Deep Learning, BNNs are of considerable interest for benchmarking performance. This analysis of training a BNN at scale outlines the limitations and benefits compared to a conventional neural network.

\end{abstract}

\section{Introduction}\label{intro}
One important challenge for machine and deep learning (DL) practitioners is to develop a robust and accurate understanding of the model uncertainty. The current state of the art deep learning networks are now able to learn representations in complex high dimensional data for doing context informed predictions. However, these predictions are often taken blindly with the provided accuracy metric, which may be erroneous. Further, for scientific applications of machine learning such as in physics, biology, and manufacturing, including accurate model uncertainties is crucial.\\  
\indent Conventional deep neural networks (DNNs) are deterministic models. These models do not provide uncertainty quantification (UQ), model confidence or a probabilistic framework for model comparison. Typically, a probabilistic model is used to compute these quantities of interest. In a deep learning context, DNNs can be integrated with probabilistic models such as Gaussian processes, which induce probability distribution over functions. A Gaussian process can be recovered from these networks in the limit of an infinite number of weights associated with probabilistic distributions (see  \cite{Neal1995,Williams96computingwith}). In a finite setting, a Bayesian Neural Network (BNN) is a DNN  with probability distributions instead of point estimates for each weight. Several foundational works on this topic such as Mackay \cite{mackay1992practical} and Neal \cite{Neal1995} have lead to BNNs gaining in popularity amongst DL practitioners. In theory these networks can overcome many limitations of DNNs such as overfitting and hyperparameter optimization but BNNs present additional workload-to-system challenges due to the increased computational costs. To address the computational complexities, techniques like variational inference (VI) are routinely applied \cite{hinton1993keeping,barber1998ensemble,Graves}. More recently methods such as stochastic VI and sampling-based VI have been developed \cite{hoffman2013stochastic}. The work in \cite{paisley2012variational,Kingma,rezende2014stochastic,titsias2014doubly,blundell2015weight} have further added thrust into the usage of BNN for a wide variety of applications such as autonomous driving control, medical diagnostics, explanatory atmospheric retrieval and uncertainty quantification of turbulence models. Further, comprehensive review on the BNN can also be looked in the work by Shridhar et.al\cite{shridhar2019comprehensive}. In addition to BNN there are a variety of other approaches for performing UQ, such as methods based on the dropout technique proposed by Hinton et.al \cite{hinton2012improving} for avoiding over fitting. This has been extended by Gal et.al \cite{gal2016dropout} as\textit{ Monte Carlo dropout} for quantifying uncertainties.\\ 
\indent Carrying out distributed training of BNNs poses several computational challenges on high performance computing (HPC) systems. Dustin et.al \cite{tran2019bayesian} outline some of the computational challenges of training BNNs at scale and show results for a 5 billion parameter “Bayesian Transformer” on 512 TPUv2 cores in machine translation and a Bayesian dynamics model for model-based planning. In Tran et.al work \cite{tran2019bayesian} they used mesh-Tensorflow \cite{shazeer2018mesh} to perform model distributed training and the authors observe linear scaling from 8-512 TPUs. Yeming et.al \cite{WenFlip} present a technique to sample the weights for each layer with minimum covariance for a BNN and report a comparison of computational performance of their techniques compared to conventional techniques on multiple CPUs. However both of these studies lack a detailed performance analysis of training a BNN benchmark. With the development of new hardware accelerators for DL, such as the massively parallel Intelligence Processing Unit (IPU) \cite{Graphcore_VI,Graphcore_MCMC} or the Wafer-Scale Engine \cite{Cerebras}, BNNs and probabilistic models are of considerable interest for benchmarking the performance of new architectures. There is a growing interest in scaling Probabilistic programming languages (PPL) such as the recent work by Baydin et.al \cite{baydin2019etalumis} who present an integrated cross-platform probabilistic execution protocol for directly coupling to existing scientific simulations. Given these efforts to analyse the  performance of BNNs/NNs on different architectures, to date, there have been  limited studies on the training performance on Intel Xeon Phi architectures \cite{viebke2019chaos}.
The main contributions of this work are as follows:
\begin{itemize}
    \item We evaluate the performance of TensorFlow \& TensorFlow Probability on the 10 PetaFlops Cray XC40 high performance supercomputer, Theta, at ALCF \cite{ALCF}. We present the training throughput using customized software builds which exploit the multi-core multi-thread system using optimized libraries such as Intel MKL, Intel MKL-DNN, Intel Numpy \cite{Intel_MKLNumpy} and Cray MPICH \cite{MPICH}. Until recently the computational overhead of BNNs together with the lack of efficient software stacks has been prohibitive to running at scale.
    \item We present a performance and scalability analysis of single node and data parallel distributed training of BNNs for two classification models, VGG-16 \cite{simonyan2014very} and Resnet-18 \cite{he2016deep}, applied to the CIFAR-10 dataset and VGG-16 applied to a 0.1 Million transformed MNIST dataset with large training batch size of 1024. 
    \item To the best of our knowledge, this work presents the first detailed analysis and measurements of BNN distributed training. We present scaling efficiencies up to 128 nodes; a comparison of times to a given accuracy between BNNs and conventional neural networks and detailed profiling results showing a breakdown of time spent in various routines.
    \item A smaller scale study on a NVIDIA-DGX station is also performed and we report the training time for the BNN models for the 0.1 Million transformed MNIST dataset up to 8 GPUs 
    \item We present an open source post-training software package \textit{BPrune} which can be used for pruning an arbitrary BNN model after training. Inference for BNNs can be slow due the Monte-Carlo sampling in each layer and we demonstrate how post-training pruning of the network is useful in deploying models in case of limited computational resources. Further, details on pruning is presented in Section \ref{sec:Pruning}. 
\end{itemize}

\indent This paper is organized as follows: we discuss the background of Bayesian neural networks in Section \ref{sec:BackBNN} and present the variational inference methods used for training the network. In Section \ref{sec:Meth} we outline the details of the BNN architecture, we also describe the dataset used in Section \ref{sec:data} for training and testing and the computational resources used for training in Section \ref{sec:software}. The results of the performed scaling study and pruning analysis is presented in Section \ref{sec:Results}. Finally, we present the discussion and conclusion in Section \ref{sec:discussion} and Section \ref{sec:Conclusion} respectively.

\section{Background on Bayesian Neural Networks}\label{sec:BackBNN}
BNNs represent the integration of a hierarchical Bayesian framework together with a neural network structure composed of recursive applications of linear weighted functions followed by nonlinear transformations. With prior distributions on weights, we are able to approximate their posterior distributions and perform posterior prediction via a variational inference framework.
Consider a model that returns a probability distribution over an output $y$ given an input $x$ and parameter $\theta$, that is $p(y|x,\theta)$. The goal is to learn these parameters from the observed data $D = \{D_i\}_{i=1}^N = \{x_i,y_i\}_{i=1}^N$. Following a Bayesian approach, we put a prior distribution $p(\theta)$ over $\theta$ and aim to obtain the posterior distribution
\[
	p(\theta|D)\propto p(\theta) p(D|\theta) = p(\theta) \prod_{i=1}^N p(y_i|\theta, x_i).
\]
\subsection{Variational Inference}
In most cases, the posterior distribution is intractable and an approximation is required. Consider using a variational family distribution $q_v(\theta)$ with parameters $v$ to approximate the posterior by minimizing the KL divergence, i.e.,
\begin{equation}
	\min_v \KL{q_v(\theta)}{p(\theta|D)}. \label{eq:minKL}
\end{equation}
where,
\begin{align*}
\KL{q_v(\theta)}{p(\theta|D)}&= \Ep{q}{\log\frac{q_v(\theta)}{p(\theta|D)}} =\Ep{q}{\log\frac{q_v(\theta) p(D)}{p(D|\theta)p(\theta)}} \\[10pt]
&= - \left\{\Ep{q}{\log p(D|\theta)}- \KL{q_v(\theta)}{p(\theta)}\right\} + \log p(D)  
\end{align*}
where the first term represents the log-likelihood of the model predictions, while the second part serves as a regularizer KL divergence between the approximate posterior $q_v(\theta)$ and the prior distribution $p(\theta)$. In the case where $\log p(D)$ is constant given prior distribution $p(\theta)$, minimizing the KL divergence in Eq. \eqref{eq:minKL} is equivalent to maximizing the objective function or the Evidence Lower BOund (ELBO) given by
\begin{equation}
L(v) = \sum_{i=1}^N \mathbb{E}_{q}{\log p(y_i|\theta,x_i)} - \beta \KL{q_v}{p}, \label{eq:Lv}
\end{equation}
where $\beta$ is a hyperparameter for tuning the degree of regularization during training of the neural network. Typically the $\beta$ hyperparameter is introduced to address the  difficulty of the {KL-term vanishing} which is often observed while training Variational Auto Encoders (VAEs). This hyperparameter was introduced by Bowman et.al \cite{bowman2015generating} as \textit{`KL-annealing'} where the $\beta$ parameter can be varied from 0 to 1 over the course of training. Extensions and modifications to the annealing procedure can be seen in \cite{higgins2017beta,alemi2017fixing}. Recently an analysis on the scheduling strategies of $\beta$ was presented by Liu et.al \cite{liu2019cyclical}. Once the BNN model is trained the inference procedure is to compute the predictive probability distribution $p(y^{*}|x^{*})$ given by
\begin{equation}
p(y^{*}|x^{*}) = \int p_{\theta}(y^{*}|x^{*})p(\theta) dw \, ,
\label{Eq:InferPost}
\end{equation}
where $x^{*}$ is the unseen (test) sample and $y^{*}$ is the corresponding predicted class.
Finding a closed form solution to the above integral for non conjugate pair of distribution is not possible hence the integral can be approximated as an expectation by sampling from $q_v(w|D)$ as
\begin{equation}
\Ep{q}{p(y^{*}|x^{*})} = \int q_v(\theta | D) p_{w}(y|x) dw \approx \frac{1}{S} \Sigma_{i=1}^{S} p_{\theta_{i}}(y^{*}|x^{*}) 
\end{equation}
where $S$ are the number of samples or Monte-Carlo iterations.
\subsection{Training Algorithm}\label{sec:TrainModel}
We summarize the procedure to optimize the loss function described in Eq. \ref{eq:Lv}. The backpropagation algorithm is at the heart of training a neural network and  relies of the calculation of gradients of the loss function which are subsequently used to update the weights layer by layer. There are several approaches available for computing the gradients such as  score-function gradient estimators shown in \cite{paisley2012variational,ranganath2013black},  reparameterization  gradient  estimators  described in \cite{blundell2015weight,Kingma}, or a combination of the two as described in Naesseth et.al \cite{naesseth2017reparameterization}. These methods provide an unbiased stochastic gradient which is used to calculate the optimum for the loss function. In the current work we use the reparametrization procedure for computing the gradients. We  use the Gaussian distribution for the variational posterior and initilize the weights, $\theta$, by sampling from the unit Gaussian with mean $\mu$ and standard-deviation $\sigma$. The hyperparameters for the variational posterior are therefore $v = (\mu, \sigma)$. The reparametization trick is to parameterize weights as a function, $t$, where $\theta = t(v, \epsilon) = \mu + {\rm{log}}(1 + exp(\sigma)) \circ \epsilon $, where $\circ$ represents an element-wise multiplication and the parameter free noise is defined by $\epsilon \sim \mathit{N}(0, I)$. Eq. \ref{eq:Lv} can be rewritten in terms of $L(\theta,v)$ and the gradients, $\Delta_{\mu} L(\theta,v)$ and $\Delta_{\sigma} L(\theta,v)$, are computed to update the hyperparamters $ \mu  \leftarrow  \mu + \eta \Delta_{\mu} L(\theta,v) $ and  $ \sigma  \leftarrow  \sigma + \eta \Delta_{\sigma} L(\theta,v) $. We refer readers to \cite{blundell2015weight} for more advanced details on back-propagation in BNNs.\\ 

\section{Methodology} \label{sec:Meth}
In Section \ref{sec:data} we discuss the network architectures and the dataset considered in this study. The details of the hardware, software, and the analysis procedure used in the study for benchmarking are discussed in Section \ref{sec:software}.

\subsection{Dataset and Network Architecture} \label{sec:data}
\begin{figure}
    \centering
    \includegraphics[scale=0.10,width=0.75\textwidth]{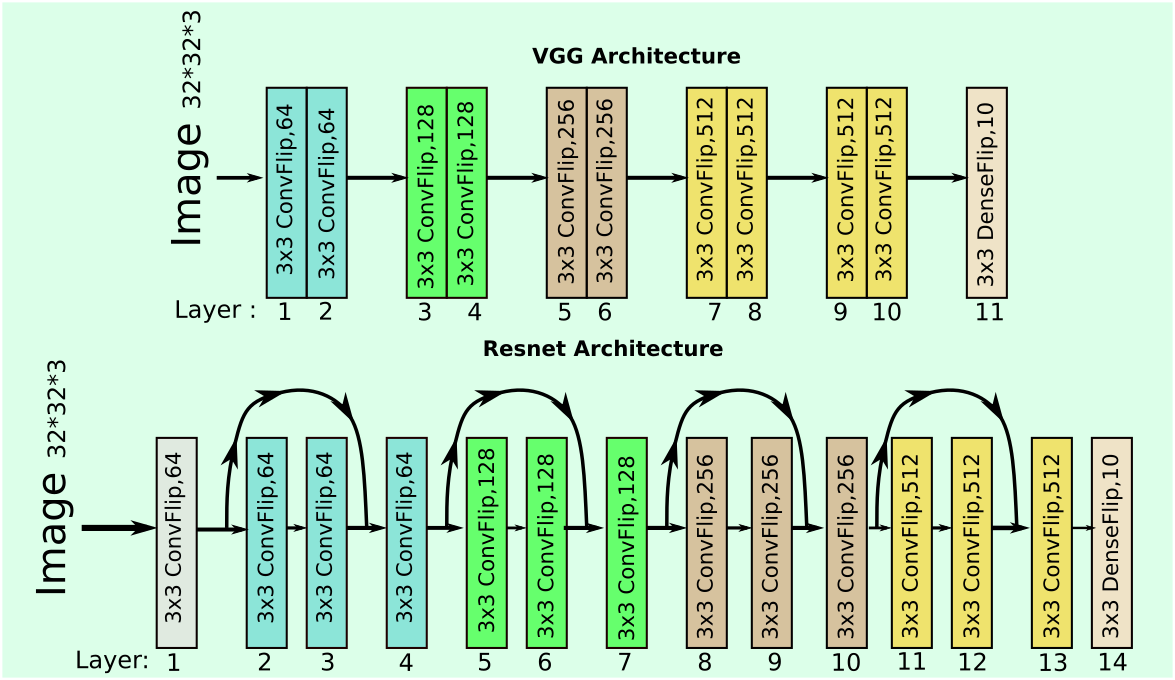}
    \caption{A schematic of the network architectures used to understand the performance of BNN. For both the VGG and Resnet architectures Batch normalization and activation layer are not shown and only Bayesian flipout convolutional layers are represented.}
    \label{Fig:NetworkArch}
\end{figure}

In this study we use the following datasets which are common in the machine learning community for classification tasks: CIFAR-10, MNIST and MNIST transformed 0.1 Million. The CIFAR-10 dataset \cite{krizhevsky2009learning} consists of 60000 32x32 colour images in 10 classes, with 6000 images per class. There are 50000 training images and 10000 test images. The MNIST dataset \cite{lecun1998mnist} consists of hand-written gray scale 28x28 images of digits representing the numbers $0-9$. There are 50,000 training and 10,000 test images in this dataset. This dataset is used for our pruning study presented in Section \ref{sec:Pruning}. The MNIST transformation 0.1 Million images were generated by pseudo-random deformations and translations of the original MNIST data using the package by Bottou et.al \cite{loosli2007training}. Given the larger volume this dataset allows us to scale up the batch size for the distributed training of the VGG-16 network.\\
 \indent In this work we demonstrate the use of distributed training of BNNs using data parallelism with Horovod \cite{sergeev2018horovod} for two image classification models, VGG-16 \cite{simonyan2014very} and Resnet-18 \cite{he2016deep}, applied to the CIFAR-10 dataset. In the conventional VGG network the architecture consists of convolutions layers with maxpooling and batch normalization operations. In the Resnet architecture the input is passed through a block consisting of convolution, batch normalization and max pooling operations. Subsequent layers in the network are structured as blocks featuring convolution, batch normalization and ReLU activations except in the last operation of a block, ReLU activation is not performed. So called `shortcut connections' are also made between the blocks with an stride of 2. Further details on the architecture and implementation can be found in He et.al \cite{he2016deep}. 
 In a BNN implementation the conventional convolutions layers are replaced by the Bayesian convolution layers which, at runtime, are sampled using a so-called \textit{Flipout technique} \cite{WenFlip}. This technique uses a sign flip operation to sample the weights with minimum covariance for each layer. The fully connected layers are also replaced by fully connected probabilitic layers. The priors are chosen to be standard Normal and VI is performed assuming the mean-field approximation \cite{blei2017variational}. The total number of trainable parameters for VGG-16 and Resnet-18 are 18 million and 9 million respectively. The representation diagram of the BNN networks are shown in Fig. \ref{Fig:NetworkArch}. For the pruning study, presented in Section \ref{sec:Pruning}, we demonstrate the use of our software package, BPrune, with a network composed of 2 Convolution Flipout and 1 Dense Flipout layers (BNN-Conv); and a second network composed of 3 Dense Flipout layers (BNN-FC). Further details about the network structures and hyperparameters can be seen in Appendix \ref{Appen:NetworkHyper}. To perform inference on a held out test set, and to estimate the prediction uncertainty intervals, we Monte Carlo sample from the approximate predictive distribution. This sampling increases the computational cost of the inference step in a BNN which can be significant in comparison to conventional neural networks. 

\subsection{Setup}\label{sec:software}
\indent A variety of software frameworks have been developed which are keeping pace with recent advances in probabilistic methods. Some of the most notable are Stan \cite{carpenter2017stan}, PyStan\cite{stan2018pystan}, Edward, Pyro\cite{bingham2019pyro}, Gen \cite{Cusumano-Towner}, Edward2 and TensorFlow Probability\cite{dillon2017tensorflow}. These frameworks are capable of running efficiently on a variety of architectures such as CPUs, GPUs and TPUs, and provide programming flexibility and ease of coding probabilistic models. Furthermore, there are various techniques for improving the scaling performance during training for conventional networks, such as pre-fetching data, the use of mixed precision and data parallel distributed training. An efficient implementation to perform data parallel training is Horovod \cite{sergeev2018horovod}, an open-source library that employs efficient inter-communications via ring reduction. \\
\indent For the current work we use Theta, a HPC cluster with 2$^{nd}$ generation Intel Xeon Phi\textsuperscript{TM} processors, code named Knights Landing (KNL), at Argonne Leadership Computing Facility. The important architectural features to note for this study are Theta's multi-core, hyper threaded nodes (64 cores per node with 4 hyper threads per core) \cite{ALCF}. We use TensorFlow (v1.14.1), TensorFlow Probability(v0.7.0) and Horovod (v0.18.1) for our benchmarking study. Tensorflow is compiled using GCC 8.2.0 and was linked to high performance math libraries such as Intel MKL and MKL-DNN libraries, while Horovod is compiled using GCC 7.3.0 and linked to the Cray MPI Library. Optimal throughput performance was observed using 1 MPI rank per node and 2 threads per core for the distributed runs. Scaling studies for the two BNN architectures are performed up to 128 nodes for CIFAR-10 keeping the regularization parameter constant as $\beta$ = 1. We also modified this parameter constant and found no improvement in the model performance.
We then use the same setup with $\beta = 1$ to  perform the scaling study of VGG-16 network for the MNIST transformed 0.1 million images. To demonstrate the effect of pruning we train two BNNs networks on the MNIST data.\\ 
\indent For performing data distributed training the Tensorflow graph scheduler allows  the order of operation execution to vary across workers, even in the case of similar models. We found that this is not helpful when performing collective operations during the distributed training of a BNN and can result into a deadlock.  Horovod v0.18.1 introduced  additional worker co-ordination logic which ensured that all workers submit collective operations in a common order. A caching scheme is implemented, where the collective operations are processed and gathered by the co-ordinating ranks only once from the worker pool. Each rank also stores the broadcasted processed results in  their cache. Horovod carries out these co-ordination processes at a frequent interval during the training which is referred to as a cycle time. Only collective operation requests are executed at the chosen cycle time across the workers. For effective network utilization Horovod has the ability to fuse individual collective operations. The procedure couples the cycle time and size of the collective messages. To find a optimal trade off between the two is challenging and therefore grouping of collective operations are carried out only at a given cycle when a complete group of requests are present. In the case of multiple complete groups they are fused into a larger message. A lower bound on fusion to complete groups indirectly leads to a minimum message size making it agnostic to the cycle time. Optimal settings here can lead to efficient network utilization. 
Further details and scaling results of the implementation can be seen in detail by Laanait et.al \cite{laanait2019exascale}. For our work we found the default cycle time of 5ms and fusion buffer size 64 MB to be optimal.

\section{Results} \label{sec:Results}
Section \ref{subsection:results:scalability} and \ref{subsection:results:efficiency} present the performance analysis and scalability results for the VGG-16 and Resnet-18 networks used in this study. The inference using BNNs are discussed in Section \ref{subsection:results:inference}. The results for the MNIST big-data run with 0.1 million images for BNN VGG-16 model is discussed in Section \ref{subsection:results:scalability2}. The Graphical Processing Unit(GPU) performance analysis results are presented in Section \ref{sec:GPUStudy}. The effect of pruning on a BNNs accuracy and a description of the BPrune pruning library are presented in Section \ref{sec:Pruning}.

\subsection{Throughput and MPI statistics}\label{subsection:results:scalability}

\begin{figure}[ht]
    \centering
    \includegraphics[scale=0.5,width=0.9\textwidth,height=\textheight,keepaspectratio]{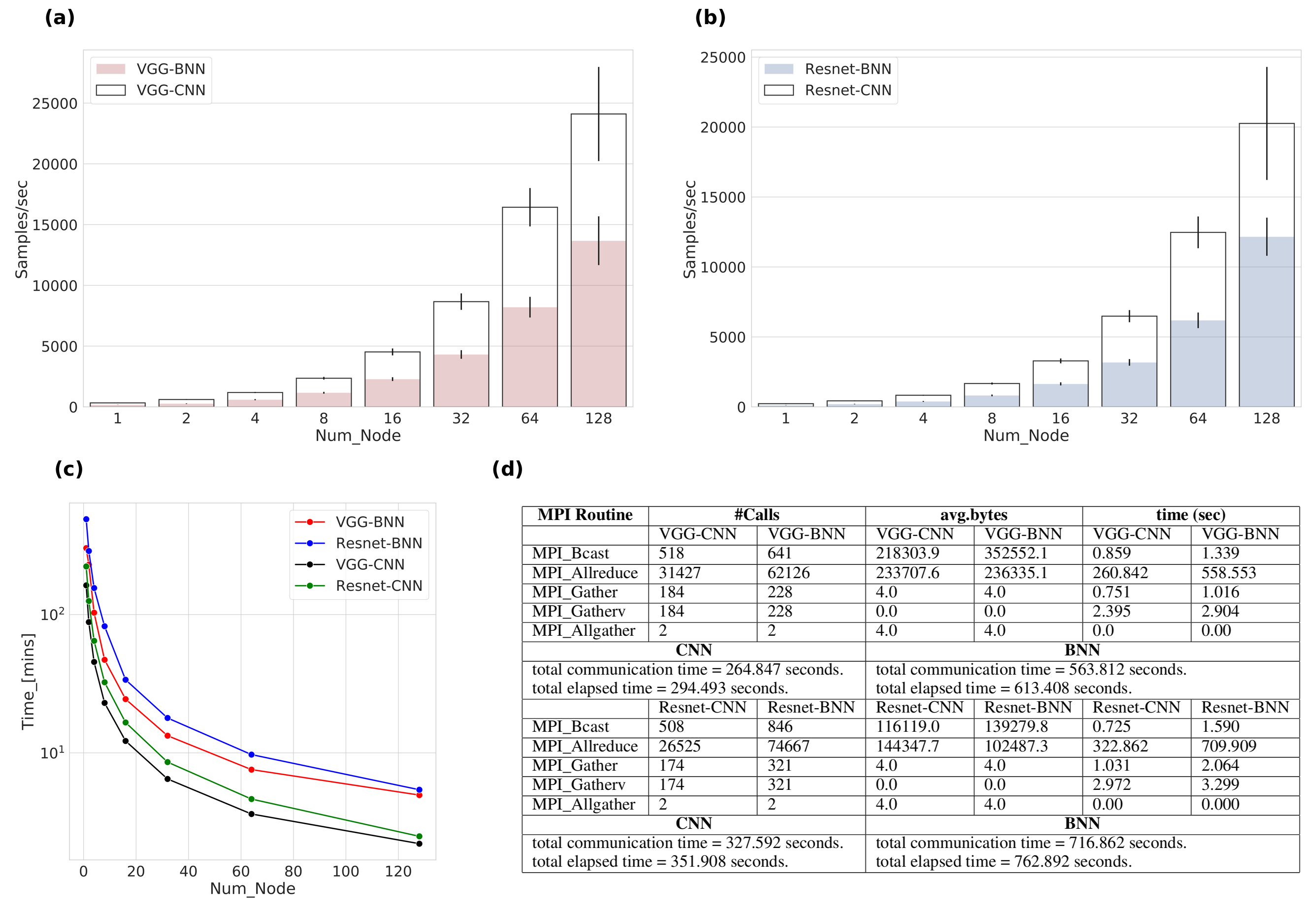}
    \caption{
    Number of samples processed per second for a BNN and CNN implementation of the VGG (a) and Resnet (b) models. (c) A comparison of the training time (mins) for a fixed number of 64 epochs with a batch size of 256 for both a BNN and CNN implementation of the VGG and Resnet models as shown in the legend. (d) Table of MPI routine statistics for rank 0 from a 64 node run generated using MPI profiling for the VGG (upper) and Resnet (lower) models.}
    \label{fig:Scaling_nodes}
\end{figure}

Figure \ref{fig:Scaling_nodes}(a) shows the measured samples processed per second verses the number of nodes for a BNN  (filled pink histogram) and conventional CNN (unfilled histogram) VGG model. A fixed batch (mini-batch) size of 128 is used. The learning rate is fixed to $1 \times 10^{-4}$ and is scaled by the number of nodes during the distributed training. The error bars represent the standard deviation of samples processed per second over all iterations. As the number of nodes used increases the samples processed per second by a BNN network is nearly 50\% less then the CNN counterpart. Similar trends are observed in the case of the Resnet architecture as shown in Fig. \ref{fig:Scaling_nodes}(b). These results can be explained by the increased computational overhead of a BNN which contains approximately double the number of trainable parameters compared to a CNN and features the flipout sampling technique for every trainable parameter in the network. The training time based on a fixed number of 64 epochs is also compared for the two architectures as shown in Fig. \ref{fig:Scaling_nodes}(c).
It can be seen that training Resnet either with a CNN or BNN takes $\sim$25\% more time in comparison to the VGG model for the single node run because of the difference in architectures. For the BNN implementation we find that as we increase the number of workers the time to train a VGG BNN on 128 nodes takes 3.37 minutes while the VGG CNN model takes 1.3 minutes; for the Resnet BNN it takes 3.3 minutes and Resnet CNN takes 1.4 minutes. Overall the BNNs are found to take approximately a factor of 2.4 increase in the time to complete a fixed number of epochs for these models. As the number of ranks increases up to 128 ranks the computation speed-up of 61\% is achieved for VGG BNN and 90\% for Resnet BNN architecture for same minibatch size.\\
\indent Table \ref{fig:Scaling_nodes}(d) shows the main MPI routines outling the number of calls, averages bytes and time in each, reported by rank 0 from a 64 node run (1 MPI rank per node) obtained from the MPI profiler HPCTW \cite{HPCTW_ALCF}. Overall the communication time for both models is significantly higher for BNNs than CNNs. In particular the all-reduce operations constitute most of the communication time and Horovod carries out more all-reduce operations for the BNN implementations of both models.

\subsection{Scaling efficiency}\label{subsection:results:efficiency}

\begin{figure}[h]
    \centering
    \includegraphics[scale=0.5,width=0.9\textwidth,height=\textheight,keepaspectratio]{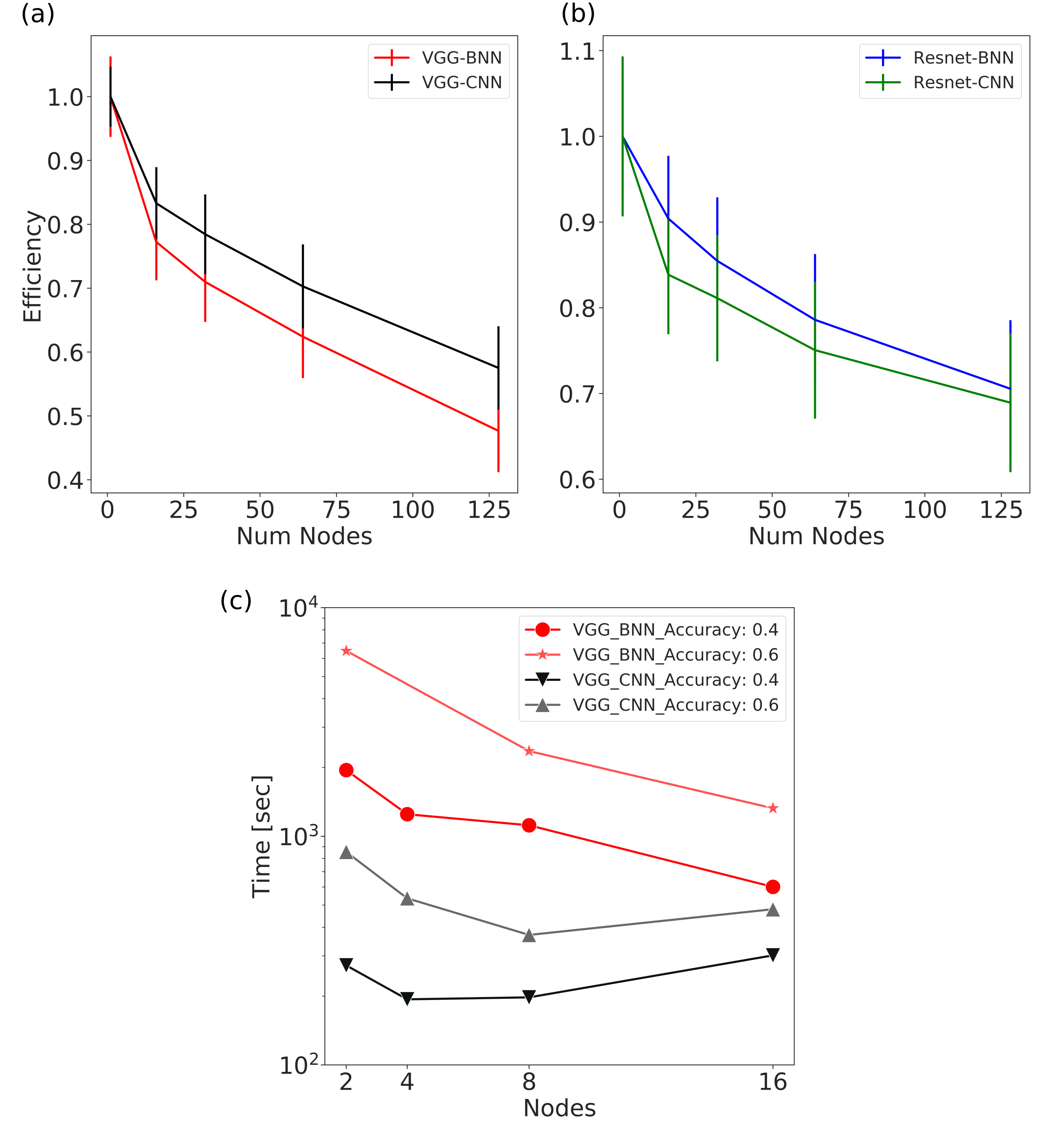}
    \caption{The scaling efficiency for the BNN and CNN implementation of VGG (a) and Resnet (c) The time taken to reach a fixed training accuracy with increasing number of nodes for the VGG-BNN and VGG-CNN architectures.}
    \label{fig:Effi_Speedup}
\end{figure}

\indent To further understand the (weak) scalability of a BNN compared to a CNN we compute the scaling efficiency as $\frac{T_{1}}{T_{N} \times N}$, where $T_{1}$ is the time to process fixed number of epochs with 1 rank, $T_{N}$ is time to process fixed number of epochs with $N$ ranks. 
Fig. \ref{fig:Effi_Speedup}(a) shows the training efficiency curve based on recorded training time shown in Fig. \ref{fig:Scaling_nodes} for VGG and Resnet with a BNN and CNN implementation. The rate of efficiency decline for the VGG BNN (red) is faster then that of the VGG CNN (black). 
We find the opposite trend for the Resnet BNN and CNN models where the BNN implementation scales better than the corresponding CNN however the two models are consistent within the 1$\sigma$ error bars. 
We find that the FLOP rate for the BNN VGG-16 model is approximately 573.87 million while for the CNN VGG it is 18.84 million. The dominant contribution for the VGG BNN model is the effect of the larger parameter set in inducing higher communication costs which makes it scale less efficiently. We record a FLOP rate of 299.02 million for the Resnet BNN model (blue) and 9.82 million for the CNN counterpart (green). The BNN implementation is slightly more compute intensive in both BNN models and Horovod operates efficiently to overlap the compute and communication. We find that the communication efficiency is higher for the BNN Resnet model compared to the CNN counterpart.\footnote{The communication efficiency is calculated as the ratio of communication time (MPI\_WTIME) to elapsed time (includes MPI\_INIT \& MPI\_FINALIZE). For 16 node run the efficiency of BNN VGG and Resnet models are 86.81\% and 87.59\% repectively, while that of the CNN VGG and Resnet model are 80.26\% and  88.59\% respectively. For a 128 node run the BNN VGG and Resnet model communication efficiencies are 91.15\% and  94.91\% while for the CNN VGG and Resnet model efficiencies are 86.99\% and 89.11\% respectively.}
Overall we find that communication costs dominate in the VGG model compared to Resnet due to the larger number of parameters in the model; which contributes to it scaling less efficiently than the Resnet model for both the CNN and BNN implementation. The time to reach a fixed training accuracy is shown Fig. \ref{fig:Effi_Speedup}(c) for the VGG BNN model. It can be seen that with the increase in the number of ranks the time to reach a fixed accuracy decreases. This plot shows the runtime for two training accuracies of 0.4  and 0.6 with different markers for each model as given in the legend. It is clear from this figure that the time to a given accuracy increases substantially for a BNN network shown in red compared to the corresponding CNN in black. We find that training the VGG model to an accuracy of 0.6 (0.4) takes approximately 7.57 (7.14) times longer for the BNN on two nodes. Running on 16 nodes we find runtimes are 2.76 (1.99) times longer for a BNN to reach an accuracy of 0.6 (0.4). Note that the minibatch size used in this case was 256. We find that when scaling up a larger effective batch helps reduce the difference between training a BNN and a CNN to the same level of accuracy.

\subsection{Inference using BNNs}\label{subsection:results:inference}

\begin{figure}[h]
    \centering
    \includegraphics[scale=0.25,width=0.85\textwidth,height=\textheight,keepaspectratio]{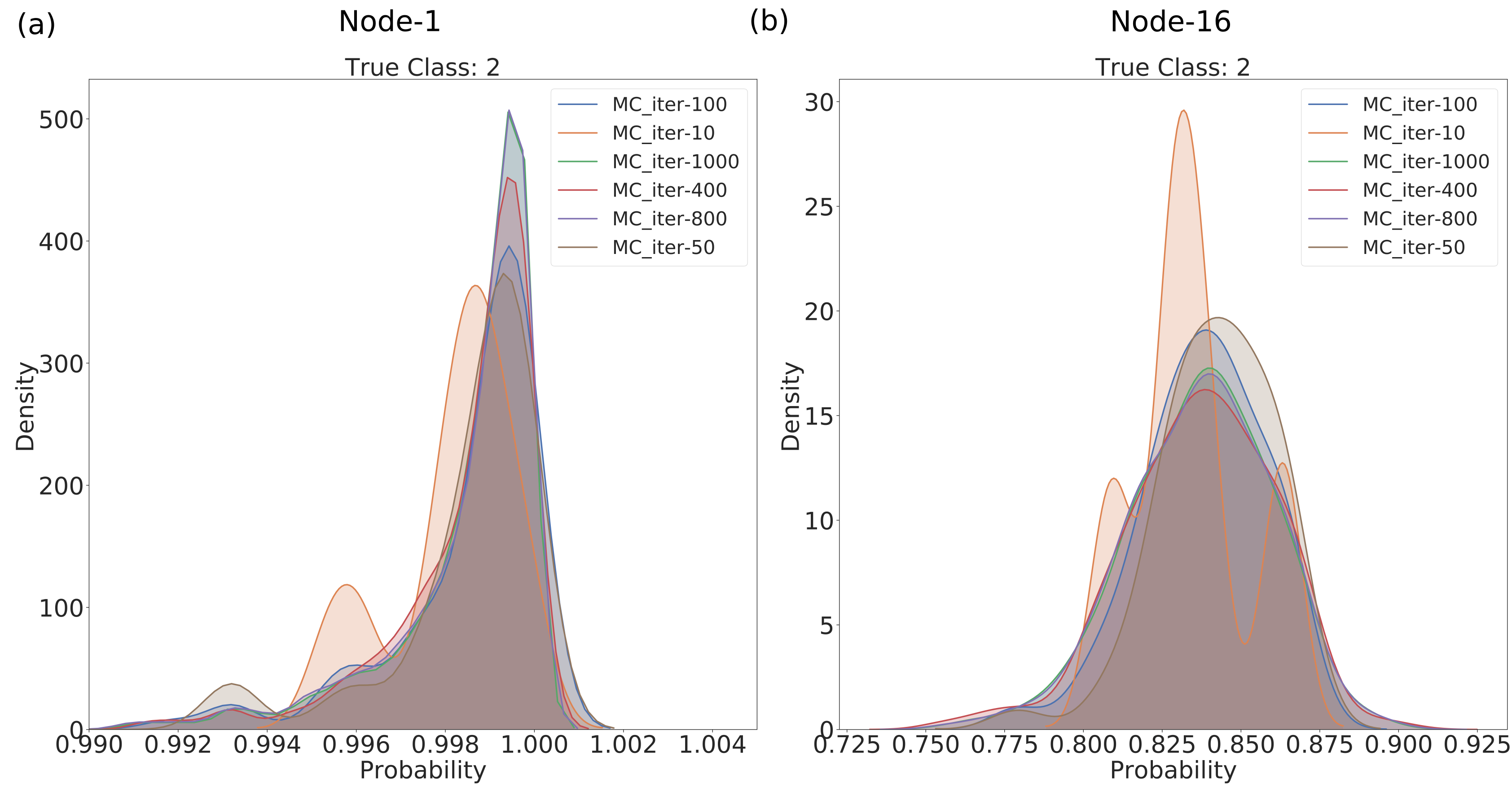}
    \caption{The BNN VGG-16 predictive pdf of softmax values for the test image to be in class 2 for a given number of MC iterations as indicated in the legend (the true image class label is 2). These are plotted using the kernel distribution estimate method (KDE) for the test CIFAR10 dataset. The Node-1 (a) and Node-16 (b) models have been trained to an accuracy of 0.91 and 0.64 respectively.}
    \label{fig:MonteCarloPosterior}
\end{figure}

\indent The BNN VGG-16 and Resnet-18 models were trained for 64 epochs on one node to a training accuracy of 0.923 and 0.942 respectively.  Using these trained models we carry out inference on the test data as in  Eq. \ref{Eq:InferPost}. To understand the effect of the number of Monte-Carlo(MC) iterations on the predictive probability density function (pdf) we vary the number of iterations and record the softmax outputs for a given true class label. In Fig. \ref{fig:MonteCarloPosterior} the predictive pdf of the softmax values for various MC number of iterations is plotted. Here we are predicting a class label of 2 for a single image when the true class label is 2. We also compare the results for the VGG BNN model trained on (a) 1 node (b) 16 nodes with training accuracies of 0.923 and 0.64 respectively. The range of MC iterations sampled from was 10-1000. From Fig.\ref{fig:MonteCarloPosterior}
it is clear that 10 MC iterations results in a noisy pdf in both panels. We find convergent results for the pdfs with MC iteration $\ge 400$. The difference in the accuracy between the Node-1 and Node-16 model is apparent here. An accuracy of 0.923 for Node-1 results in a narrow pdf close to 1 for MC iterations $\ge$ 400 while the reduced accuracy of the Node-16 model results in a pdf with larger variance. Note that we see similar trends for other test images. The runtime for inference using  10 MC iterations was $\sim$5\% of the runtime for 1000 iterations, which were found to be 152.4 seconds and 2861.0 seconds respectively. Clearly the number of MC iterations plays an important role in the computational cost of inference in a BNN. \\

\begin{figure}
    \centering
    \includegraphics[scale=1.0,width=0.9\textwidth,height=\textheight,keepaspectratio]{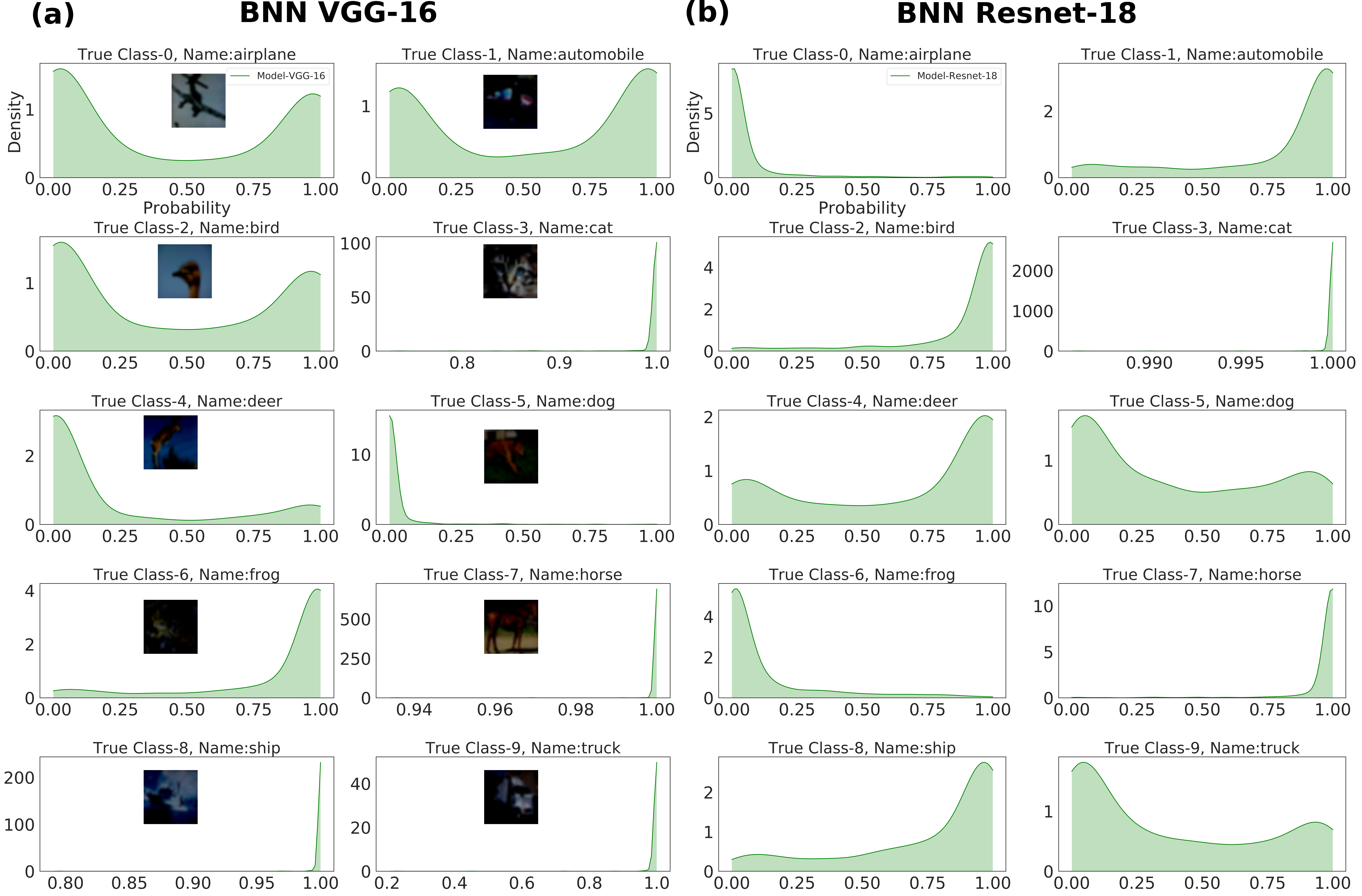}
    \caption{The predictive pdf  of the softmax values output from the VGG-16 model in (a) and the Resnet model in (b) using a sample of test images in each class. These  distributions represent the softmax values for class X when the true class is X (where X is in the range 0-9 as in the plot titles) and are obtained by running the trained BNN model for 400 MC iterations.} 
    \label{fig:TestdataPosterior}
\end{figure}

\indent In Fig. \ref{fig:TestdataPosterior} we show the predictive pdf of the softmax values output from the VGG-16 model in (a) and the Resnet model in (b) using a sample of test images in each class. These  distributions represent the softmax values for class X when the true class is X and are obtained by running the trained BNN model for 400 MC iterations. In this figure if the model was performing perfectly on the test dataset each panel would show a sharp peak at unity.  Note that both models were trained on a single node for a  fixed number of epochs to a training accuracy of 0.923 and 0.942 for VGG and Resnet respectively. We include these results as representative outputs from two different model implementations of a Bayesian neural network applied to the same test set. 

Comparing the results from each model on the same test image it can be seen that for some of the images both models behave similarly  correctly identifing the image, as for test class 7 and 3. 
For test image labels 0, 4 and 5 when one model precisely misclassifies the image (sharp peak around zero probability) the other model shows large uncertainties in their classification.  For test image label 6 the VGG model correctly classifies with little variance while the Resnet model misclassifies with little variance. Overall the output from a BNN contains a richer set of information compared to a corresponding CNN, which can be used to understand model performance and carry out model comparisons. We shall explore  this topic in future work.

\subsection{Scalability of VGG-16 on MNIST-Transformed}\label{subsection:results:scalability2}
\begin{figure}
    \centering
    \includegraphics[scale=0.5,width=0.95\linewidth]{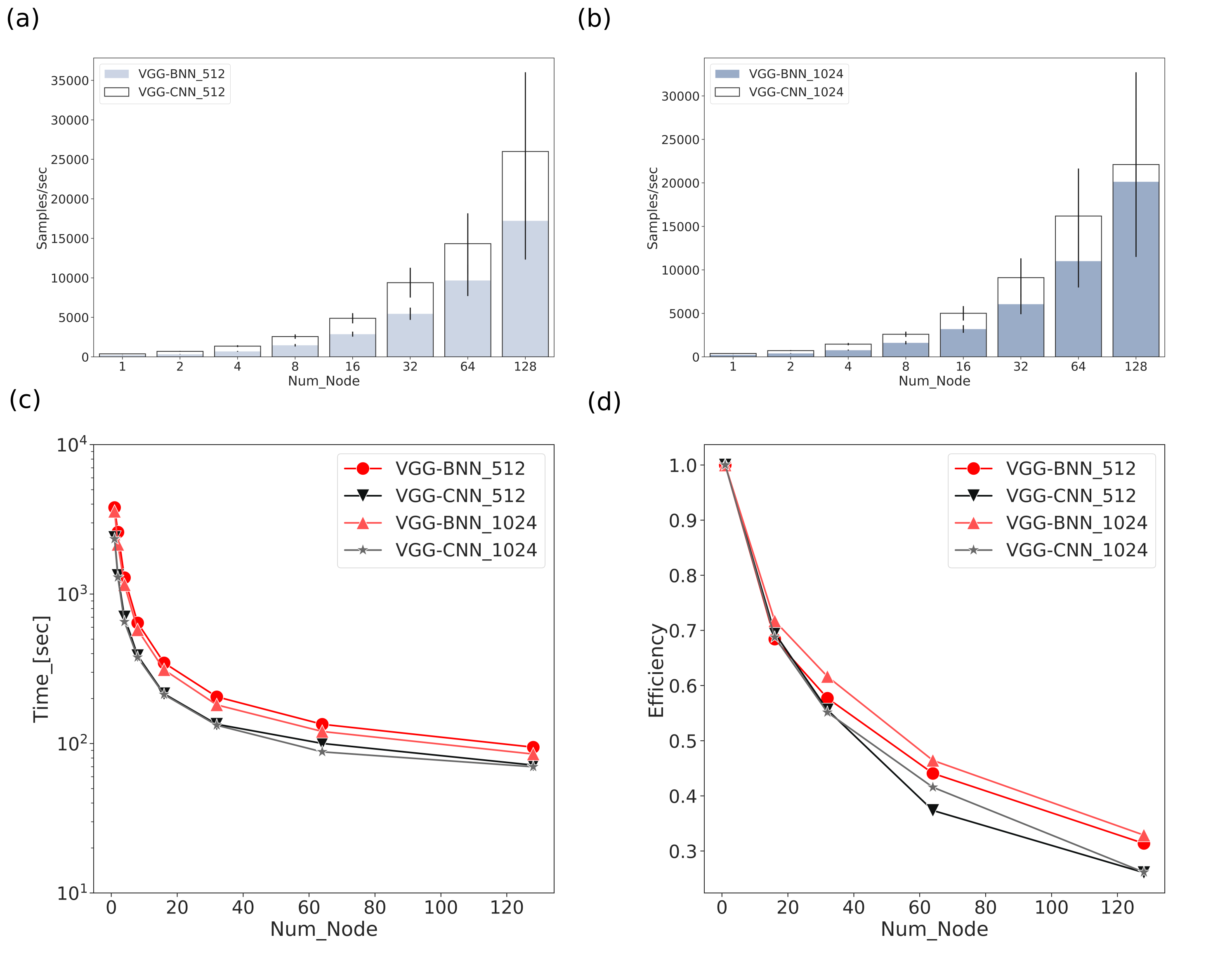}
    \caption{The measured samples per second processed by the BNN VGG-16 model versus the number of nodes for the MNIST transformed 0.1M images for a batch size of 512 (a) and  1024 (b). (c) The training time in seconds with the increasing number of nodes for different batch-size run for 12 epochs. 
    (d) The efficiency curve for BNN, CNN VGG model with two different batch sizes with increasing number of nodes.}
    \label{fig:mnist0p1M_result}
\end{figure}

HPC resources at leadership computing facilities are typically used to process and train DL models with big data. Therefore, we extended the distributed training performance evaluation study of BNNs  using the VGG-16  model applied to the 0.1 million MNIST transformed image data-set \cite{loosli2007training}. When scaling up the number of nodes the number of epochs was fixed to 12 and all other hyperparameters and priors are unchanged from the previous study.  
Fig.\ref{fig:mnist0p1M_result} shows a histogram of the measured  samples processed per second versus the number of nodes for the VGG model with a batch size of 512 in panel (a) and 1024 in panel (b). The error bars (black) represent the standard deviation in samples per second over all iterations. As the batch size increases from 512 to 1024 we see a change in the mean sample per second processed in the range of 5-12\% for every node from 1-128 for VGG BNN model.
From this figure it is clear that increasing the throughput for the CNN model saturates at a batchsize of 512 with little improvement using 1024 but the throughput for the BNN does improve using the larger batch size. 
Figure \ref{fig:mnist0p1M_result}(c) shows the total runtime to complete 12 epochs for 512 and 1024 minibatch sizes using the CNN and BNN models. As expected using larger minbatchs reduces the time to complete a fixed number of epochs with similar improvement going from 512 to 1024 minibatch size in the CNN and BNN. For example using 16 nodes the runtime is reduced by $\sim$10\% using a minibatch size of 1024 compared to 512.
Making a qualitative comparison with Fig. \ref{fig:Scaling_nodes}(c) it is clear that increasing the batch size as we increase the number of nodes improves the performance of the BNN model and can significantly reduce the difference in training time when comparing with the corresponding CNN.
This effect of the larger minibatch size can also be seen in the efficiency plot shown in Fig. \ref{fig:mnist0p1M_result}(d).
In this figure the larger minbatch size reduces the difference in scaling efficiency between the BNN and CNN.
However the actual efficiencies for 512 and 1024 minibatch sizes are much lower at 128 nodes compared to those in Fig. \ref{fig:Effi_Speedup}. 

\subsection{GPU Scaling Study}\label{sec:GPUStudy}
\begin{figure}
    \centering
    \includegraphics[scale=0.7, width=0.95\textwidth,keepaspectratio]{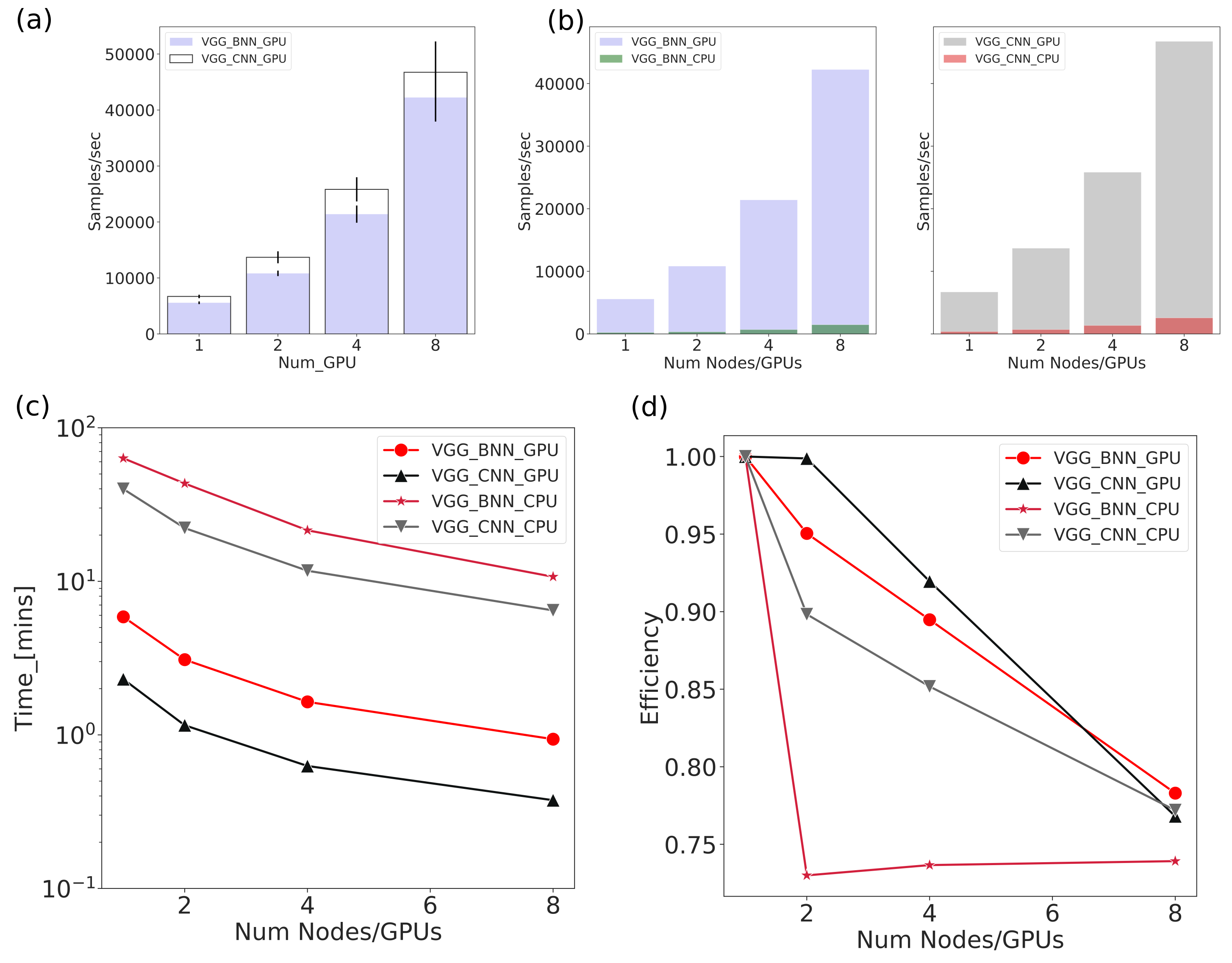}
    \caption{The VGG-BNN scaling performance on a NVIDIA-DGX V100 work station for the MNIST 0.1M image dataset with a minibatch size of 512. (a) The samples processed per second with increasing number of GPUs. Panels (b) compared the throughput on GPUs and CPUs for the BNN and CNN VGG models. (c) The total runtime to complete 12 epochs with increasing number of GPUs and CPUs for CNN and BNN model. (d) The efficiency of both the BNN and CNN models with increasing number of GPUs and CPUs. }
    \label{fig:GPU_Scaling}
\end{figure}

We analyze the performance of the VGG-16 BNN and CNN networks on the NVIDIA-DGX-1 V100 GPU cluster\footnote{The configuration for the GPUs used at ALCF is as follows, 8X Tesla V100, GPU total system memory of 128 GB, a CPU with Dual 20-Core Intel Xeon E5-2698 v4 2.2 GHz, 40,960 NVIDIA CUDA Cores, 5,120 NVIDIA Tensor Cores, System memory of 512 GB 2,133 MHz DDR4 LRDIMM, Storage of 4X 1.92 TB SSD RAID 0, and Dual 10 GbE, 4 IB EDR network} at ALCF. 
The model training was performed with a minbatch sizes of 512 and 1024 using the Adam optimizer and a learning rate of $1 \times 10^{-4}$. The MNIST Transformed dataset is used for training up to 8 GPUs for  12 epochs. All other distributed training settings are the same as in previous sections for comparison. Fig. \ref{fig:GPU_Scaling} (a) shows the image samples processed per second verses the number of GPUs. It can be seen that the throughput for both the BNN and CNN scales linearly with the increase in the number of GPUs.
Fig. \ref{fig:GPU_Scaling}(b) shows a comparison between the throughput for the VGG BNN model on the GPU (shaded blue histrogram) and previous results on the CPU (green filled histogram).
Fig. \ref{fig:GPU_Scaling}(c) shows a similar comparison for the CNN model on the GPU adn CPU.
It is clear from the figure that the throughput capabilites of GPUs is signifcantly higher for both the BNN and CNN model. Using 8 GPUs we find $\sim$29 times more samples per second compared to running on the same number of Xeon KNL nodes for BNN model and $\sim$18.20 times more samples per second for the CNN model on the GPU compared to the CPU.
For both the BNN and CNN models we find that nearly 128 Xeon Phi nodes are needed to achieve the same throughput as 4 V100 GPUs. Overall, in the study 8 GPUs gave the best training throughput for a batchsize of 512, with $\sim$42K samples for BNN and $\sim$ 46K samples for CNN.\\

\indent The runtime for 12 epochs for the VGG model on the CPUs and GPUs are shown in Fig.\ref{fig:GPU_Scaling}(c). We find that the difference in runtime for the BNN is $\sim 10$x less on a single GPU compared to the CPU and $\sim$11x less on 8 GPUs compared to 8 CPUs.As a stark comparison we find that the runtime for 12 epochs on 8 GPUs is 1.67 times faster than using 128 KNL nodes for the BNN model highlighting the difference in performance on the different architectures.As shown in previous sections the runtime for the BNN model is larger than the corresponding CNN regardless of whether we are using a CPU or GPU. On 8 GPUs the run time of the BNN is 56.21 sec while the CNN runtime is 22.51 sec; on 8 CPU nodes the runtimes for the BNN and CNN models are 641.79 sec and 387.58 sec respectively.
Fig. \ref{fig:GPU_Scaling}(d) shows the efficiency curve for the VGG BNN and CNN models. The scaling trends are similar between the BNN and CNN up to 8 GPUs. We find that the efficiency scaling is better on the GPUs compared to the CPUs with less discrepancy between the BNN and CNN. In future work we will examine the scaling behaviour to higher numbers of GPUs.

\subsection{Post training pruning of BNNs}\label{sec:Pruning}
The process of removing weights from a neural network is referred as pruning. This procedure has been introduced in the past to reduce the network complexity and improve generalization \cite{lecun1990optimal,giles1994pruning}. In the case of BNNs with probabilistic layers the procedure has been found to be helpful in limiting the computational cost and memory demands. The study by Graves \cite{Graves} have found that pruning can also improve the final performance by reducing the noise in the gradient estimates. In addition, the work by Blundell et.al\cite{blundell2015weight} has shown in a BNN pruning experiment that by reducing a given network by up to 95\% the accuracy is not significantly affected. For various practical applications of BNN models, for example when deployed for inference on edge devices, they will bring with them a higher computational cost. This overhead is due to the number of MC iterations required to produce robust predictions and uncertainty estimates. As shown in Fig. \ref{fig:MonteCarloPosterior} with the  VGG-16 BNN network, if someone chooses to run 1000 MC iterations it is 2.5 times more computationally expensive in comparison to 400 iteration and 18.3 times in comparison to 10 iterations. With a pruned smaller network with sparse computations the inference time can be reduced for BNNs without sacrificing accuracy.

\indent In this work we present an open source software package, {\it{BPrune}}\footnote{The github account will be open source following publication.}, which has been developed as an add-on for Tensorflow and Tensorflow Probability. This software automates the post training pruning procedure using a user defined signal-to-noise threshold to zero out weights from the graph.
The signal to noise threshold is set by the user. Although BPrune is intended for use  to prune a trained network prior to carrying out inference, it should be installed before training as information on the final graph needs to be extracted and saved.
BPrune brings with it the necessary utilities to write the required files  after training as well as to reload the trained model for inference or prunning.
A high level description of how BPrune operates is as follows:
\begin{itemize}
   \item Once training of the BNN is completed the model can be saved in the TensorFlow native file format which writes \textit{model-$\textless$iteration$\textgreater$.ckpt} in the checkpoint directory. This binary file contains the trained weight matrices of the BNN. This file is output using the conventional procedure of Tensorflow  \cite{tensorflow2015-whitepaper}.
   \item TensorFlow probability lacks routines to easily reload a saved model for restarts, inference or for any additional model modifications such as pruning. BPrune addresses this by outputting  two additional  text files, \textit{LayerNames.txt} and \textit{OpsNames.txt}, after training.
   \item \textit{LayerNames.txt}: This file contains an ordered list of Bayesian and Non-Bayesian Layer names (also known as training Variables) associated with each layer. This is needed for accessing each layers respective weight matrices. For example using a Gaussian prior for the weights, each layer in the network will have two matrices defined in the name scope of the layer, i.e a matrix for mean ($\mu$) and sigma ($\sigma$) named as `kernel' and  `un-transformed scale' respectively. These names are listed into the file associated with each layer. At inference time BPrune parses this file and identifies the layers associated with training variable for pruning.
   \item \textit{OpsNames.txt}: This file contains all the operations which are defined by Tensorflow to execute the graph. This information is needed during the re-load to identify the input and output placeholders, the names of the sampling operation and other metrics such as accuracy, if defined. 
   For example, in the case of a BNN model trained to perform image classification we define a categorical distribution for the prediction output and compute the log probability. This operation is identified as `log\_prob' for a label distribution name scope and is listed as one of the operations in the OpsNames.txt file. BPrune parses the file to automatically identify the minimum required operations for running inference or pruning with the test/heldout dataset.  
   \item To run the model for the inference and pruning the user needs to provide the case directory including the two text files and the checkpoint directory containing the saved checkpoint files, to BPrune. If a user wishes to prune the model a threshold value can be passed as a command line argument which has a default set to 10.0. The user can also set the number of MC iterations to be used for generating prediction samples.
   \item When pruning BPrune  automatically calculates the signal-to-noise(SNR)\footnote{The ratio for the Gaussian prior over weights can be simply calculated as a ratio  $|\mu| / \sigma$ \cite{Graves}. In the initial BPrune release  Gaussian priors are supported. Other choices of distribution will be supported in future releases.} ratio for individual layers and zeros the weights below the given threshold.
   \item BPrune outputs a binary file which contains the used test samples, the labels, the total number of non zeros in the network, the non zeros per layer, the predictive probability distributions and the inference runtime. 
\end{itemize}
The motivation for creating the BPrune package is to address the limitation of re-loading the trained Tensorflow Probability based model for inference and  provide a framework for  pruning a BNN model by computing the SNR for each layer and then automatically pruning them based on a user defined threshold. BPrune also computes the signal to noise ratio for the complete network by collecting the individual layer ratio as a single array to visualize the global signal to noise ratio for the whole network, saving manual intervention and efforts. The code is agnostic to loading model graphs either trained serially or using distributed training. A user can load the trained model at any iteration by simply specifying the name of the Tensorflow checkpoint file (`model-$\textless$iteration$\textgreater$.ckpt'). In addition, the object oriented programming approach used for developing the framework can be used to run multiple independent inference or pruning jobs using high throughput computing in parallel.   

\begin{figure}
    \centering
    \includegraphics[scale=0.001,width=0.95\textwidth,keepaspectratio]{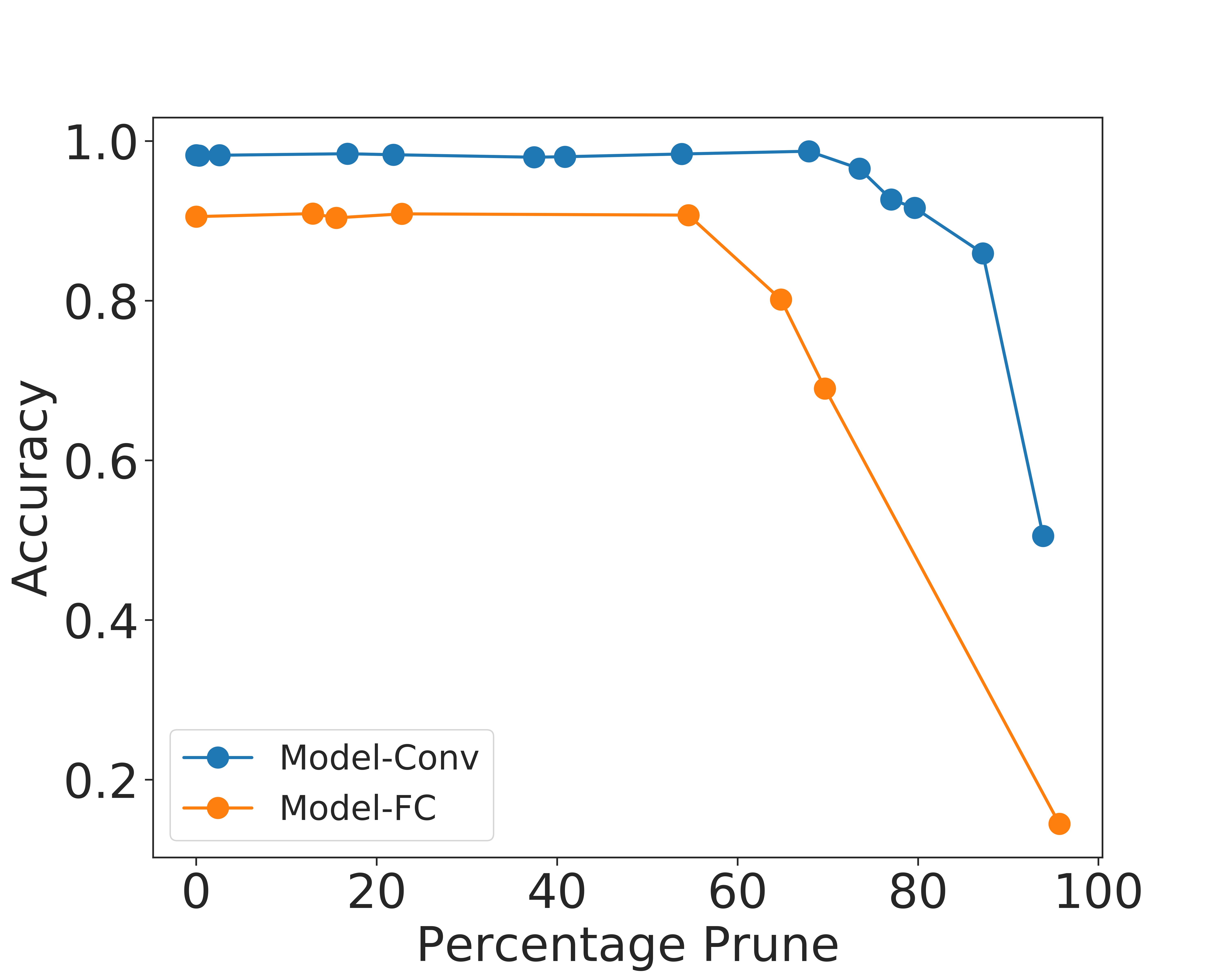}
    \caption{The test accuracy for Model-Conv and Model-FC versus varying percentage of the network pruned (higher percentages indicate high sparsity in the network) using the MNIST test data.}
    \label{fig:PruningVsAccuracy}
\end{figure}

\begin{figure}
    \centering
    \includegraphics[scale=0.25,width=0.95\textwidth,height=\textheight,keepaspectratio]{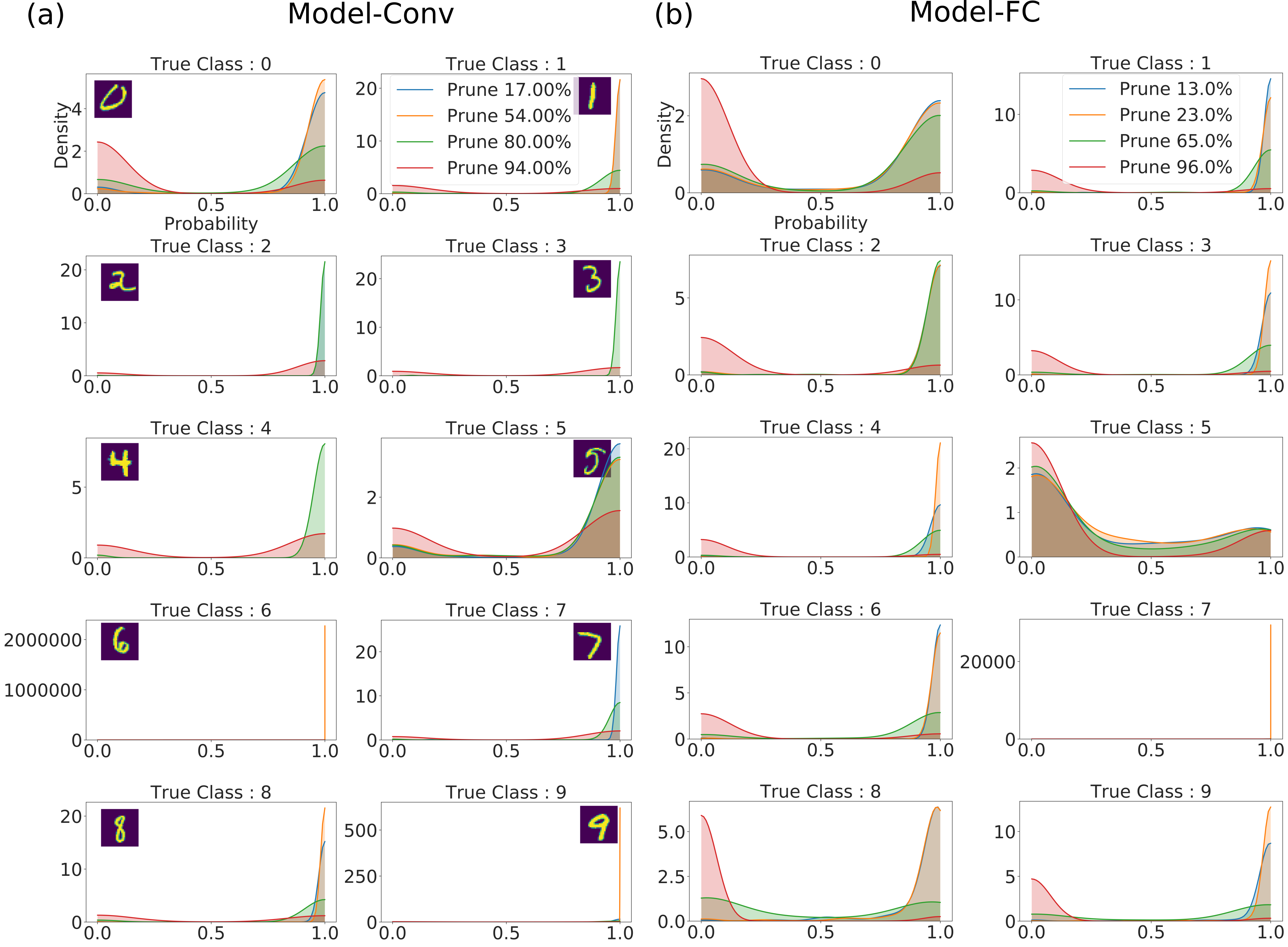}
    \caption{The predictive pdfs for the softmax values from 200 MC iterations for a sample of test images generated by running the trained models Model-Conv (a) and Model-FC (b) for inference with differing percentages of pruning as given in the legend. }
    \label{fig:Testdata_posteriorPrune}
\end{figure}

\indent To demonstrate BPrune in action we train two BNNs on the MNIST dataset. One network has convolutional and fully connected layers while the second network has only fully connected layers. These models will be referred to as Model-Conv and Model-FC respectively.
 The details of these network architectures and training parameters can be found in Appendix \ref{Appen:NetworkHyper}.
 Both Model-Conv and Model-FC are trained for 23 epochs until the training accuracy is 0.98 and 0.90 respectively. We then output each trained model to a Tensorflow native file format with the full graph and parameter information needed by BPrune. In Fig. \ref{fig:PruningVsAccuracy} we show the effect on the test accuracy when varying the signal-to-noise threshold of weights pruned from the network. This corresponds to a pruning percentage as indicated on the x-axis in Fig. \ref{fig:PruningVsAccuracy} i.e retaining all weights regardless of their signal-to-noise would represent zero pruning, while removing all weights below a fixed signal-to-noise threshold corresponds to a certain percentage of the network being removed. For the particular models used here we find that the Model-Conv test accuracy remains unchanged as the pruning percentage increases to approximately 80\% of the network. The Model-FC test accuracy is unchanged after pruning to approximately 60\% of the network and additional pruning after this reduces the accuracy. To illustrate the effect of pruning on inference, we profiled the Model-FC Bayesian network inference with 0\% and 70\% of the model pruned for a single MC iteration. We find that the inference time is reduced by $\sim$50\% for the 70\% pruned network compared to running the full network.
 It is worth noting that the current Tensorflow graph operations do not support sparse computation which may  speed-up matrix multiplication in the pruned network. Introducing this functionality in future could significantly speed-up the inference for pruned networks and will be explored in future work.\\
 In addition it is worth considering the effect of pruning on the prediction uncertainties for individual classes. We show the predictive pdfs for the softmax values for a sample of test images generated by running the trained models Model-Conv (a) and Model-FC (b) for inference with 200 MC iterations with differing percentages of pruning in Fig. \ref{fig:Testdata_posteriorPrune}. Note that in this plot the pdf of softmax values for a class X from 200 MC iterations are shown when the actual true class is X, for the MNIST dataset. Remarkably for some of the test images chosen, pruning to over 90\% of the network has little impact on the predictive pdf as seen for Model-Conv for class 3, 6, 8 and 9; and for Model-FC for class 7. For both models, the test image for the digit 5 we can see that the model looses the confidence in the prediction as the network is pruned. For this image with $\sim$0\% pruning we see a peaked pdf at unity while at $\sim$90\% pruning the pdf becomes bimodel with peaks at 0 and 1 showing the uncertainty caused by losing relevant weights from the network. Comparing the effect of pruning on the Model-Conv and Model-FC pdfs for these chosen test images, we find that pruning affects the Model-FC network to a greater extent that Model-Conv as can be seen for test image 8 and 9 in Fig. \ref{fig:Testdata_posteriorPrune}. Convolutional networks have proven to be significantly more robust for computer vision applications compared to fully connected networks alone and these results are consistent with our results or pruning BNNs.

\section{Discussion}\label{sec:discussion}
The training and inference results presented in this work used a customized build of the distributed training framework, Horovod, which implements data-parallel training. On Theta the MPI communication between the Horovod workers uses Cray MPICH libraries. Horovod copies the Tensorflow operation graph across different ranks to process unique mini-batches of training data. To compute the average gradient from each rank an all-reduce operation is inserted at the back-propagation computation.\\

\indent During this study we used Horovod versions 0.16 and 0.18.  Using version 0.16 we encountered failures when running the BNN models on $\ge$ 4 nodes as some ranks were stalling due to a failure in negotiating all reduce. On further analysis, we found that the  stalls and failures using version 0.16 could be  attributed to the following breakdown in Horovod operations: (a) Rank 0 is the central scheduler in Horovod operations and waits on each rank to accumulate tensors for broadcasting and all-reduce. (b) As the Tensorflow process is independent to schedule its operations in the graph, if one or more ranks experience load imbalance and attempts to execute the all-reduce operation in different orders the operation will stall resulting in a deadlock. We found that the BNN model suffered from this load imbalance which was causing the failures.\\
\indent Horovod version 0.18 \cite{sergeev2018horovod} features some significant improvements in how the all reduce is carried out amongst the workers. In this version the developers have introduced a dynamical reordering of all-reduce operations so that consistency can be maintained for all ranks. This is called negotiating all-reduce. The is done by improving the gradient reduction strategy. The worker co-ordination is done by a light weight BitAllRedude and grouping of gradient tensors are performed based on the buffer size to perform reduction operations for the buffered tensors at fixed interval (cycle-time). The implementation provides a improvement in various aspects of running deep learning at scale on large models and big data with Horovod. In profiling the BNNs in this work, we found that the network has over a thousand all-reduce operations per step which results in the  controller being forced to  receive and then send millions of messages per second for larger jobs. This bottleneck for BNN models was reduced by improving the collective reduction implementation. Our findings outline the challenges in running BNNs at scale due to the large reductions operations in conjunction with more compute,which require efficient use of hardware and systems interconnect together with optimize MPI routines.\\

\section{Conclusions}\label{sec:Conclusion}
We present a performance analysis for the distributed training of Bayesian neural networks. As the use of machine learning for decision making increases for scientific and industrial applications, quantifying uncertainties becomes increasingly important. BNNs are compute intensive and in a distributed setting requires significant communication; as a result training large BNN models present unique challenges for future architectures.
The following are the main results of this paper for the image classification models VGG-16 and Resnet-18:
\begin{itemize}
    \item The throughput for BNNs are approximately 50\% less then the corresponding CNN for small batch sizes on KNL nodes.
    \item For BNN small batch sizes we find approximately a factor of 2.4 increase in the runtime for a fixed number of epochs.
    \item The neural network model size plays a role in the scalibility and efficiency with increasing number of nodes and BNNs can suffer from reduced efficiency as we scale up. BNN efficiency can be improved to match or outperform the corresponding CNN by increasing the batch size or varying the Horovod cycle time and buffer size.
    \item Overall we see a 30x increase in the FLOP rate for BNNs compared to CNNs.
    \item Runtime to a fixed accuracy can be up to a factor of $\sim 7\times$ longer on a small number of nodes but reduced to a factor of $\sim 3\times$ longer on $\ge$ 16 nodes.
    \item Increasing the batch size to 512 or 1024 can improve the BNN's performance and reduce the difference in training time and throughtput compared to the corresponding CNN, especially as the number of nodes increases.
    \item For inference a minimum of 400 MC iterations is needed to produce robust pdfs.
    \item Using 8 GPUs, we find a $\sim 29 (18)\times$ increase in the throughput for BNNs (CNNs) compared to running on an equivalent number of KNL nodes.
    \item For certain architectures pruning to 60\% of the network has little effect on the inference accuracy. The effect of pruning the majority of the network e.g. 90\% has noticeable effects of the final pdf produced for certain test images.
\end{itemize}

Our findings indicate that the computational overheads of training a BNN network can be reduced with a distributed training framework which employs efficient MPI communication schemes. The reported results on the Cray XC40 could also be used as a baseline for projections on the next generation of hardware architectures.
We hope that this study is useful to the data science community providing insights into the distributed training with Horovod on HPC clusters. 
We plan to extend this work to understand the the scalability of BNNs with different models such as Recurrent Neural Networks, Autoencoders and Generative Adversarial Networks with large datasets.

\section{Acknowledgements}
This research used resources of the Argonne Leadership Computing Facility, which is a DOE Office of Science User Facility supported under Contract DE-AC02-06CH11357. This research was funded in part and used resources of the Argonne Leadership Computing Facility, which is a DOE Office of Science User Facility supported under Contract DE-AC02-06CH11357. This paper describes objective technical results and analysis. Any subjective views or opinions that might be expressed in the paper do not necessarily represent the views of the U.S. DOE or the United States Government. Declaration of Interests - None.

\bibliographystyle{unsrt}

\appendix{

\section{ Network Information \& Hyper-parameter details} \label{Appen:NetworkHyper}
Information about the VGG-16 and  Resnet-18 networks together with details of the individual layers is shown in Table \ref{tab:VGGArch} and Table \ref{tab:Resnet-table} respectively. For  distributed training the learning rate was scaled by the number of nodes. The initial learning rate was fixed to $1 \times 10^{-4}$ with batch size of 256 and Relu activation function. A mean field normal distribution with zero mean and unit variance is used to define the prior $p(\theta)$ for the weights, the approximate kernel posterior $q(\theta |x)$ is initialized by normal distribution with mean an -9.0 and standard deviation as 0.1. The distributed model training were carried using the Adam optimizer. \\
\indent The models  used for the pruning study are shown in Table \ref{tab:BNn_pruneConv}-\ref{tab:BNNFCPrune} with Relu as the activation function in each layer. Model training was done serially, with a learning rate of $1 \times 10^{-3}$ and a batch size of 100 with RMSProp as the optimizer. 

\begin{table}[]
\caption{BNN VGG-16 Architecture used for distributed training}
\label{tab:VGGArch}
\begin{tabular}{lll}
\hline
\textbf{Layer (type)}                                                                                                                  & \textbf{Output Shape} & \textbf{Param \#} \\ \hline
input\_1 (InputLayer)                                                                                                                  & (None, 32, 32, 3)     & 0                 \\ \hline
ConV\_I\_0 (Conv2DFlipout)                                                                                                             & (None, 32, 32, 64)    & 3520              \\ \hline
batch\_normalization\_v1                                                                                                               & (None, 32, 32, 64)    & 256               \\ \hline
activation (Activation)                                                                                                                & (None, 32, 32, 64)    & 0                 \\ \hline
ConV\_II\_0 (Conv2DFlipout)                                                                                                            & (None, 32, 32, 64)    & 73792             \\ \hline
batch\_normalization\_v1\_1                                                                                                            & (None, 32, 32, 64)    & 256               \\ \hline
activation\_1 (Activation)                                                                                                             & (None, 32, 32, 64)    & 0                 \\ \hline
Max\_I\_0 (MaxPooling2D)                                                                                                               & (None, 16, 16, 64)    & 0                 \\ \hline
ConV\_I\_1 (Conv2DFlipout)                                                                                                             & (None, 16, 16, 128)   & 147584            \\ \hline
batch\_normalization\_v1\_2                                                                                                            & (None, 16, 16, 128)   & 512               \\ \hline
activation\_2 (Activation)                                                                                                             & (None, 16, 16, 128)   & 0                 \\ \hline
ConV\_II\_1 (Conv2DFlipout)                                                                                                            & (None, 16, 16, 128)   & 295040            \\ \hline
batch\_normalization\_v1\_3                                                                                                            & (None, 16, 16, 128)   & 512               \\ \hline
activation\_3 (Activation)                                                                                                             & (None, 16, 16, 128)   & 0                 \\ \hline
Max\_I\_1 (MaxPooling2D)                                                                                                               & (None, 8, 8, 128)     & 0                 \\ \hline
ConV\_I\_2 (Conv2DFlipout)                                                                                                             & (None, 8, 8, 256)     & 590080            \\ \hline
batch\_normalization\_v1\_4                                                                                                            & (None, 8, 8, 256)     & 1024              \\ \hline
activation\_4 (Activation)                                                                                                             & (None, 8, 8, 256)     & 0                 \\ \hline
ConV\_II\_2 (Conv2DFlipout)                                                                                                            & (None, 8, 8, 256)     & 1179904           \\ \hline
batch\_normalization\_v1\_5                                                                                                            & (None, 8, 8, 256)     & 1024              \\ \hline
activation\_5 (Activation)                                                                                                             & (None, 8, 8, 256)     & 0                 \\ \hline
Max\_I\_2 (MaxPooling2D)                                                                                                               & (None, 4, 4, 256)     & 0                 \\ \hline
ConV\_I\_3 (Conv2DFlipout)                                                                                                             & (None, 4, 4, 512)     & 2359808           \\ \hline
batch\_normalization\_v1\_6                                                                                                            & (None, 4, 4, 512)     & 2048              \\ \hline
activation\_6 (Activation)                                                                                                             & (None, 4, 4, 512)     & 0                 \\ \hline
ConV\_II\_3 (Conv2DFlipout)                                                                                                            & (None, 4, 4, 512)     & 4719104           \\ \hline
batch\_normalization\_v1\_7                                                                                                            & (None, 4, 4, 512)     & 2048              \\ \hline
activation\_7 (Activation)                                                                                                             & (None, 4, 4, 512)     & 0                 \\ \hline
Max\_I\_3 (MaxPooling2D)                                                                                                               & (None, 2, 2, 512)     & 0                 \\ \hline
ConV\_I\_4 (Conv2DFlipout)                                                                                                             & (None, 2, 2, 512)     & 4719104           \\ \hline
batch\_normalization\_v1\_8                                                                                                            & (None, 2, 2, 512)     & 2048              \\ \hline
activation\_8 (Activation)                                                                                                             & (None, 2, 2, 512)     & 0                 \\ \hline
ConV\_II\_4 (Conv2DFlipout)                                                                                                            & (None, 2, 2, 512)     & 4719104           \\ \hline
batch\_normalization\_v1\_9                                                                                                            & (None, 2, 2, 512)     & 2048              \\ \hline
activation\_9 (Activation)                                                                                                             & (None, 2, 2, 512)     & 0                 \\ \hline
Max\_I\_4 (MaxPooling2D)                                                                                                               & (None, 1, 1, 512)     & 0                 \\ \hline
flatten (Flatten)                                                                                                                      & (None, 512)           & 0                 \\ \hline
Dense\_I\_4 (DenseFlipout)                                                                                                             & (None, 10)            & 10250             \\ \hline
\textbf{\begin{tabular}[c]{@{}l@{}}Total params: 18,829,066\\ Trainable params: 18,823,178\\ Non-trainable params: 5,888\end{tabular}} &                       &                   \\ \hline
\end{tabular}
\end{table}

\begin{table}[]
\caption{CNN VGG-16 Architecture used for distributed training}
\label{tab:CNNVGGArch}
\begin{tabular}{lll}
\hline
\multicolumn{1}{c}{\textbf{Layer (type)}}                                                                                 & \multicolumn{1}{c}{\textbf{Output Shape}} & \multicolumn{1}{c}{\textbf{Param \#}} \\ \hline
input\_1 (InputLayer)                                                                                                       & (None, 32, 32, 3)                          & 0                                      \\ \hline
ConV\_I\_0 (Conv2D)                                                                                                         & (None, 32, 32, 64)                         & 1792                                   \\ \hline
batch\_normalization (BatchNo                                                                                               & (None, 32, 32, 64)                         & 256                                    \\ \hline
activation (Activation)                                                                                                     & (None, 32, 32, 64)                         & 0                                      \\ \hline
ConV\_II\_0 (Conv2D)                                                                                                        & (None, 32, 32, 64)                         & 36928                                  \\ \hline
batch\_normalization\_1 (Batch                                                                                              & (None, 32, 32, 64)                         & 256                                    \\ \hline
activation\_1 (Activation)                                                                                                  & (None, 32, 32, 64)                         & 0                                      \\ \hline
Max\_I\_0 (MaxPooling2D)                                                                                                    & (None, 16, 16, 128)                        & 0                                      \\ \hline
ConV\_I\_1 (Conv2D)                                                                                                         & (None, 16, 16, 128)                        & 73856                                  \\ \hline
batch\_normalization\_2 (Batch                                                                                              & (None, 16, 16, 128)                        & 512                                    \\ \hline
activation\_2 (Activation)                                                                                                  & (None, 16, 16, 128)                        & 0                                      \\ \hline
ConV\_II\_1 (Conv2D)                                                                                                        & (None, 16, 16, 128)                        & 147584                                 \\ \hline
batch\_normalization\_3 (Batch                                                                                              & (None, 16, 16, 128)                        & 512                                    \\ \hline
activation\_3 (Activation)                                                                                                  & (None, 16, 16, 128)                        & 0                                      \\ \hline
Max\_I\_1 (MaxPooling2D)                                                                                                    & (None, 8, 8, 256)                          & 0                                      \\ \hline
ConV\_I\_2 (Conv2D)                                                                                                         & (None, 8, 8, 256)                          & 295168                                 \\ \hline
batch\_normalization\_4 (Batch                                                                                              & (None, 8, 8, 256)                          & 1024                                   \\ \hline
activation\_4 (Activation)                                                                                                  & (None, 8, 8, 256)                          & 0                                      \\ \hline
ConV\_II\_2 (Conv2D)                                                                                                        & (None, 8, 8, 256)                          & 590080                                 \\ \hline
batch\_normalization\_5 (Batch                                                                                              & (None, 8, 8, 256)                          & 1024                                   \\ \hline
activation\_5 (Activation)                                                                                                  & (None, 8, 8, 256)                          & 0                                      \\ \hline
Max\_I\_2 (MaxPooling2D)                                                                                                    & (None, 4, 4, 512)                          & 0                                      \\ \hline
ConV\_I\_3 (Conv2D)                                                                                                         & (None, 4, 4, 512)                          & 1180160                                \\ \hline
batch\_normalization\_6 (Batch                                                                                              & (None, 4, 4, 512)                          & 2048                                   \\ \hline
activation\_6 (Activation)                                                                                                  & (None, 4, 4, 512)                          & 0                                      \\ \hline
ConV\_II\_3 (Conv2D)                                                                                                        & (None, 4, 4, 512)                          & 2359808                                \\ \hline
batch\_normalization\_7 (Batch                                                                                              & (None, 4, 4, 512)                          & 2048                                   \\ \hline
activation\_7 (Activation)                                                                                                  & (None, 4, 4, 512)                          & 0                                      \\ \hline
Max\_I\_3 (MaxPooling2D)                                                                                                    & (None, 2, 2, 512)                          & 0                                      \\ \hline
ConV\_I\_4 (Conv2D)                                                                                                         & (None, 2, 2, 512)                          & 2359808                                \\ \hline
batch\_normalization\_8 (Batch                                                                                              & (None, 2, 2, 512)                          & 2048                                   \\ \hline
activation\_8 (Activation)                                                                                                  & (None, 2, 2, 512)                          & 0                                      \\ \hline
ConV\_II\_4 (Conv2D)                                                                                                        & (None, 2, 2, 512)                          & 2359808                                \\ \hline
batch\_normalization\_9 (Batch                                                                                              & (None, 2, 2, 512)                          & 2048                                   \\ \hline
activation\_9 (Activation)                                                                                                  & (None, 2, 2, 512)                          & 0                                      \\ \hline
Max\_I\_4 (MaxPooling2D)                                                                                                    & (None, 1, 1, 512)                          & 0                                      \\ \hline
flatten (Flatten)                                                                                                           & (None, 512)                                & 0                                      \\ \hline
Dense\_I\_4 (Dense)                                                                                                         & (None,10)                                  & 5130                                   \\ \hline
\begin{tabular}[c]{@{}l@{}}\textbf{Total params: 9,421,898}\\\textbf{ Trainable params: 9,416,010}\\ \textbf{Non-trainable params: 5,888}\end{tabular} &                                            &                                        \\ \hline
\end{tabular}
\end{table}

\begin{table}[]
\caption{BNN Resnet-18 Architecture used for distributed training}
\label{tab:Resnet-table}
\begin{tabular}{llll}
\hline
\textbf{Layer (type)}                                                                                                       & \textbf{Output Shape} & \textbf{Param \#} & \textbf{Connected to}                                                                                         \\ \hline
input\_1 (InputLayer)                                                                                                       & (None,32,32,3)        & 0                 &                                                                                                               \\ \hline
cond2d\_flipout                                                                                                             & (None,32,32,64)       & 3520              & input\_1{[}0{]}{[}0{]}                                                                                        \\ \hline
batch\_normalization                                                                                                        & (None,32,32,64)       & 256               & cond2d\_flipout{[}0{]}{[}0{]}                                                                                 \\ \hline
activation                                                                                                                  & (None,32,32,64)       & 0                 & batch\_normalization{[}0{]}{[}0{]}                                                                            \\ \hline
conv2d\_flipout\_2                                                                                                          & (None,32,32,64)       & 73792             & activation{[}0{]}{[}0{]}                                                                                      \\ \hline
batch\_normalization\_1                                                                                                     & (None,32,32,64)       & 256               & conv2d\_flipout\_2{[}0{]}{[}0{]}                                                                              \\ \hline
activation\_1                                                                                                               & (None,32,32,64)       & 0                 & batch\_normalization\_1{[}0{]}{[}0{]}                                                                         \\ \hline
conv2d\_flipout\_3                                                                                                          & (None,32,32,64)       & 73792             & activation\_1{[}0{]}{[}0{]}                                                                                   \\ \hline
conv2d\_flipout\_1                                                                                                          & (None,32,32,64)       & 8256              & activation{[}0{]}{[}0{]}                                                                                      \\ \hline
add                                                                                                                         & (None,32,32,64)       & 0                 & \begin{tabular}[c]{@{}l@{}}cond2d\_flipout\_3{[}0{]}{[}0{]}\\ cond2d\_flipout\_1{[}0{]}{[}0{]}\end{tabular}   \\ \hline
batch\_normalization\_2                                                                                                     & (None,32,32,64)       & 256               & add{[}0{]}{[}0{]}                                                                                             \\ \hline
activation\_2                                                                                                               & (None,32,32,64)       & 0                 & batch\_normalization\_2{[}0{]}{[}0{]}                                                                         \\ \hline
conv2d\_flipout\_5                                                                                                          & (None,16,16,128)      & 147584            & activation\_2{[}0{]}{[}0{]}                                                                                   \\ \hline
batch\_normalization\_3                                                                                                     & (None,16,16,128)      & 512               & conv2d\_flipout\_5{[}0{]}{[}0{]}                                                                              \\ \hline
activation\_3                                                                                                               & (None,16,16,128)      & 0                 & batch\_normalization\_3{[}0{]}{[}0{]}                                                                         \\ \hline
conv2d\_flipout\_6                                                                                                          & (None,16,16,128)      & 295040            & activation\_3{[}0{]}{[}0{]}                                                                                   \\ \hline
conv2d\_flipout\_4                                                                                                          & (None,16,16,128)      & 16512             & activation\_2{[}0{]}{[}0{]}                                                                                   \\ \hline
add\_1                                                                                                                      & (None,16,16,128)      & 0                 & \begin{tabular}[c]{@{}l@{}}cond2d\_flipout\_6{[}0{]}{[}0{]}\\ cond2d\_flipout\_4{[}0{]}{[}0{]}\end{tabular}   \\ \hline
batch\_normalization\_4                                                                                                     & (None,16,16,128)      & 512               & add\_1{[}0{]}{[}0{]}                                                                                          \\ \hline
activation\_4                                                                                                               & (None,16,16,128)      & 0                 & batch\_normalization\_4{[}0{]}{[}0{]}                                                                         \\ \hline
conv2d\_flipout\_8                                                                                                          & (None,8,8,256)        & 590080            & activation\_4{[}0{]}{[}0{]}                                                                                   \\ \hline
batch\_normalization\_5                                                                                                     & (None,8,8,256)        & 1024              & conv2d\_flipout\_8{[}0{]}{[}0{]}                                                                              \\ \hline
activation\_5                                                                                                               & (None,8,8,256)        & 0                 & batch\_normalization\_5{[}0{]}{[}0{]}                                                                         \\ \hline
conv2d\_flipout\_9                                                                                                          & (None,8,8,256)        & 1179904           & activation\_5{[}0{]}{[}0{]}                                                                                   \\ \hline
conv2d\_flipout\_7                                                                                                          & (None,8,8,256)        & 65792             & activation\_4{[}0{]}{[}0{]}                                                                                   \\ \hline
add\_2                                                                                                                      & (None,8,8,256)        & 0                 & \begin{tabular}[c]{@{}l@{}}cond2d\_flipout\_9{[}0{]}{[}0{]}\\ cond2d\_flipout\_7{[}0{]}{[}0{]}\end{tabular}   \\ \hline
batch\_normalization\_6                                                                                                     & (None,8,8,256)        & 1024              & add\_2{[}0{]}{[}0{]}                                                                                          \\ \hline
activation\_6                                                                                                               & (None,8,8,256)        & 0                 & batch\_normalization\_6{[}0{]}{[}0{]}                                                                         \\ \hline
conv2d\_flipout\_11                                                                                                         & (None,4,4,512)        & 2359808           & activation\_6{[}0{]}{[}0{]}                                                                                   \\ \hline
batch\_normalization\_7                                                                                                     & (None,4,4,512)        & 2048              & conv2d\_flipout\_11{[}0{]}{[}0{]}                                                                             \\ \hline
activation\_7                                                                                                               & (None,4,4,512)        & 0                 & batch\_normalization\_7{[}0{]}{[}0{]}                                                                         \\ \hline
conv2d\_flipout\_12                                                                                                         & (None,4,4,512)        & 4719104           & activation\_7{[}0{]}{[}0{]}                                                                                   \\ \hline
conv2d\_flipout\_10                                                                                                         & (None,4,4,512)        & 262656            & activation\_6{[}0{]}{[}0{]}                                                                                   \\ \hline
add\_3                                                                                                                      & (None,4,4,512)        & 0                 & \begin{tabular}[c]{@{}l@{}}cond2d\_flipout\_12{[}0{]}{[}0{]}\\ cond2d\_flipout\_10{[}0{]}{[}0{]}\end{tabular} \\ \hline
batch\_normalization\_8                                                                                                     & (None,4,4,512)        & 2048              & add\_3{[}0{]}{[}0{]}                                                                                          \\ \hline
activation\_8                                                                                                               & (None,4,4,512)        & 0                 & batch\_normalization\_8{[}0{]}{[}0{]}                                                                         \\ \hline
average\_pooling2d                                                                                                          & (None,1,1,512)        & 0                 & activation\_8{[}0{]}{[}0{]}                                                                                   \\ \hline
flatten                                                                                                                     & (None,512)            & 0                 & average\_pooling2d{[}0{]}{[}0{]}                                                                              \\ \hline
dense\_flipout                                                                                                              & (None,10)             & 10250             & flatten{[}0{]}{[}0{]}                                                                                         \\ \hline
\begin{tabular}[c]{@{}l@{}}\textbf{Total params: 9,814,026}\\ \textbf{Trainable params: 9,810,058} \\ \textbf{Non-trainable params: 3,968 }\end{tabular} & \multicolumn{3}{l}{}                                                                                                                                     \\ \hline
\end{tabular}
\end{table}

\begin{table}[]
\caption{CNN Resnet-18 Architecture used for distributed training}
\label{tab:CNNResnetModel}
\begin{tabular}{llll}
\hline
\multicolumn{1}{c}{\textbf{Layer (type)}}                                                                                 & \multicolumn{1}{c}{\textbf{Output Shape}} & \multicolumn{1}{c}{\textbf{Param \#}} & \textbf{Connected to}                                                                                     \\ \hline
input\_1 (InputLayer)                                                                                                       & (None, 32, 32, 3)                          & 0                                      &                                                                                                           \\ \hline
Conv\_block\_I (Conv2D)                                                                                                     & (None, 32, 32, 64)                         & 1792                                   & input\_1{[}0{]}{[}0{]}                                                                                    \\ \hline
batch\_normalization (BatchNorma                                                                                            & (None, 32, 32, 64)                         & 256                                    & Conv\_block\_I{[}0{]}{[}0{]}                                                                              \\ \hline
activation (Activation)                                                                                                     & (None, 32, 32, 64)                         & 0                                      & batch\_normalization{[}0{]}{[}0{]}                                                                        \\ \hline
Conv\_block\_I\_0 (Conv2D)                                                                                                  & (None, 32, 32, 64)                         & 36928                                  & activation{[}0{]}{[}0{]}                                                                                  \\ \hline
batch\_normalization\_1 (BatchNor                                                                                           & (None, 32, 32, 64)                         & 256                                    & Conv\_block\_I\_0{[}0{]}{[}0{]}                                                                           \\ \hline
activation\_1 (Activation)                                                                                                  & (None, 32, 32, 64)                         & 0                                      & batch\_normalization\_1{[}0{]}{[}0{]}                                                                     \\ \hline
Conv\_block\_II\_0 (Conv2D)                                                                                                 & (None, 32, 32, 64)                         & 36928                                  & activation\_1{[}0{]}{[}0{]}                                                                               \\ \hline
conv2d (Conv2D)                                                                                                             & (None, 32, 32, 64)                         & 4160                                   & activation{[}0{]}{[}0{]}                                                                                  \\ \hline
add (Add)                                                                                                                   & (None, 32, 32, 64)                         & 0                                      & \begin{tabular}[c]{@{}l@{}}Conv\_block\_II\_0{[}0{]}{[}0{]}\\ conv2d{[}0{]}{[}0{]}\end{tabular}      \\ \hline
batch\_normalization\_2 (BatchNor                                                                                           & (None, 32, 32, 64)                         & 256                                    & add{[}0{]}{[}0{]}                                                                                         \\ \hline
activation\_2 (Activation)                                                                                                  & (None, 32, 32, 64)                         & 0                                      & batch\_normalization\_2{[}0{]}{[}0{]}                                                                     \\ \hline
Conv\_block\_I\_1 (Conv2D)                                                                                                  & (None, 16, 16, 128)                        & 73856                                  & activation\_2{[}0{]}{[}0{]}                                                                               \\ \hline
batch\_normalization\_3 (BatchNor                                                                                           & (None, 16, 16, 128)                        & 512                                    & Conv\_block\_I\_1{[}0{]}{[}0{]}                                                                           \\ \hline
activation\_3 (Activation)                                                                                                  & (None, 16, 16, 128)                        & 0                                      & batch\_normalization\_3{[}0{]}{[}0{]}                                                                     \\ \hline
Conv\_block\_II\_1 (Conv2D)                                                                                                 & (None, 16, 16, 128)                        & 147584                                 & activation\_3{[}0{]}{[}0{]}                                                                               \\ \hline
conv2d\_1 (Conv2D)                                                                                                          & (None, 16, 16, 128)                        & 8320                                   & activation\_2{[}0{]}{[}0{]}                                                                               \\ \hline
add\_1 (Add)                                                                                                                & (None, 16, 16, 128)                        & 0                                      & \begin{tabular}[c]{@{}l@{}}Conv\_block\_II\_1{[}0{]}{[}0{]}\\ conv2d\_1{[}0{]}{[}0{]}\end{tabular} \\ \hline
batch\_normalization\_4 (BatchNor                                                                                           & (None, 16, 16, 128)                        & 512                                    & add\_1{[}0{]}{[}0{]}                                                                                      \\ \hline
activation\_4 (Activation)                                                                                                  & (None, 16, 16, 128)                        & 0                                      & batch\_normalization\_4{[}0{]}{[}0{]}                                                                     \\ \hline
Conv\_block\_I\_2 (Conv2D)                                                                                                  & (None, 8, 8, 256)                          & 295168                                 & activation\_4{[}0{]}{[}0{]}                                                                               \\ \hline
batch\_normalization\_5 (BatchNor                                                                                           & (None, 8, 8, 256)                          & 1024                                   & Conv\_block\_I\_2{[}0{]}{[}0{]}                                                                           \\ \hline
activation\_5 (Activation)                                                                                                  & (None, 8, 8, 256)                          & 0                                      & batch\_normalization\_5{[}0{]}{[}0{]}                                                                     \\ \hline
Conv\_block\_II\_2 (Conv2D)                                                                                                 & (None, 8, 8, 256)                          & 590080                                 & activation\_5{[}0{]}{[}0{]}                                                                               \\ \hline
conv2d\_2 (Conv2D)                                                                                                          & (None, 8, 8, 256)                          & 33024                                  & activation\_4{[}0{]}{[}0{]}                                                                               \\ \hline
add\_2 (Add)                                                                                                                & (None, 8, 8, 256)                          & 0                                      & \begin{tabular}[c]{@{}l@{}}Conv\_block\_II\_2{[}0{]}{[}0{]}\\ conv2d\_2{[}0{]}{[}0{]}\end{tabular}    \\ \hline
batch\_normalization\_6 (BatchNor                                                                                           & (None, 8, 8, 256)                          & 1024                                   & add\_2{[}0{]}{[}0{]}                                                                                      \\ \hline
activation\_6 (Activation)                                                                                                  & (None, 8, 8, 256)                          & 0                                      & batch\_normalization\_6{[}0{]}{[}0{]}                                                                     \\ \hline
Conv\_block\_I\_3 (Conv2D)                                                                                                  & (None, 4, 4, 512)                          & 1180160                                & activation\_6{[}0{]}{[}0{]}                                                                               \\ \hline
batch\_normalization\_7 (BatchNor                                                                                           & (None, 4, 4, 512)                          & 2048                                   & Conv\_block\_I\_3{[}0{]}{[}0{]}                                                                           \\ \hline
activation\_7 (Activation)                                                                                                  & (None, 4, 4, 512)                          & 0                                      & batch\_normalization\_7{[}0{]}{[}0{]}                                                                     \\ \hline
Conv\_block\_II\_3 (Conv2D)                                                                                                 & (None, 4, 4, 512)                          & 2359808                                & activation\_7{[}0{]}{[}0{]}                                                                               \\ \hline
conv2d\_3 (Conv2D)                                                                                                          & (None, 4, 4, 512)                          & 131584                                 & activation\_6{[}0{]}{[}0{]}                                                                               \\ \hline
add\_3 (Add)                                                                                                                & (None, 4, 4, 512)                          & 0                                      & \begin{tabular}[c]{@{}l@{}}Conv\_block\_II\_3{[}0{]}{[}0{]}\\ conv2d\_3{[}0{]}{[}0{]}\end{tabular}        \\ \hline
batch\_normalization\_8 (BatchNor                                                                                           & (None, 4, 4, 512)                          & 2048                                   & add\_3{[}0{]}{[}0{]}                                                                                      \\ \hline
activation\_8 (Activation)                                                                                                  & (None, 4, 4, 512)                          & 0                                      & batch\_normalization\_8{[}0{]}{[}0{]}                                                                     \\ \hline
average\_pooling2d (AveragePooli                                                                                            & (None, 4, 4, 512)                          & 0                                      & activation\_8{[}0{]}{[}0{]}                                                                               \\ \hline
flatten (Flatten)                                                                                                           & (None, 512)                                & 0                                      & average\_pooling2d{[}0{]}{[}0{]}                                                                          \\ \hline
Dense\_I\_3 (Dense)                                                                                                         & (None, 10)                                 & 5130                                   & flatten{[}0{]}{[}0{]}                                                                                     \\ \hline
\begin{tabular}[c]{@{}l@{}}
\textbf{Total params: 4,912,458}\\
\textbf{Trainable params: 4,908,490}\\
\textbf{Non-trainable params: 3,968}\\ \end{tabular} &                                            &                                        &                                                                                                           \\ \hline
\end{tabular}
\end{table}

\begin{table}[h]
\caption{The BNN-Conv model used to demonstrate pruning with BPrune }
\label{tab:BNn_pruneConv}
\begin{tabular}{lll}
\hline
\textbf{Layer (type)}                                                                                                            & \textbf{Output Shape} & \textbf{Param \#} \\ \hline
input\_1 (InputLayer)                                                                                                            & (None, 28, 28, 1)     & 0                 \\ \hline
Conv\_1 (Conv2DFlipout)                                                                                                          & (None, 5, 5, 256)     & 13056             \\ \hline
Max\_I\_1 (MaxPooling2D)                                                                                                         & (None, 2, 2, 256)     & 0                 \\ \hline
Conv\_2 (Conv2DFlipout)                                                                                                          & (None, 5, 5, 256)     & 3277056           \\ \hline
flatten (Flatten)                                                                                                                & (None, 2560)          & 0                 \\ \hline
Dense\_I\_4 (DenseFlipout)                                                                                                       & (None, 10)            & 512010            \\ \hline
\textbf{\begin{tabular}[c]{@{}l@{}}Total params: 3,802,122\\ Trainable params: 3,802,122\\ Non-trainable params: 0\end{tabular}} & \multicolumn{2}{l}{}                     \\ \hline
\end{tabular}
\end{table}

\begin{table}[]
\caption{The BNN-FC model used to demonstrate pruning with BPrune }
\label{tab:BNNFCPrune}
\begin{tabular}{lll}
\hline
\textbf{Layer (type)}                                                                                                        & \textbf{Output Shape} & \textbf{Param \#} \\ \hline
input\_1 (InputLayer)                                                                                                        & (None, 784,)          & 0                 \\ \hline
den\_1 (DenseFlipout)                                                                                                        & (None, 256)           & 401664            \\ \hline
den\_2 (DenseFlipout)                                                                                                        & (None, 256)           & 131328            \\ \hline
den\_3 (DenseFlipout)                                                                                                        & (None, 10)            & 5130              \\ \hline
\textbf{\begin{tabular}[c]{@{}l@{}}Total params: 538,122\\ Trainable params: 538,122\\ Non-trainable params: 0\end{tabular}} & \multicolumn{2}{l}{}                     \\ \hline
\end{tabular}
\end{table}

}


\end{document}